\documentclass[journal]{IEEEtran}

\ifCLASSINFOpdf
   \usepackage[pdftex]{graphicx}
   \graphicspath{{../pdf/}{../jpeg/}}
   \DeclareGraphicsExtensions{.pdf,.jpeg,.png}
\else
   \usepackage[dvips]{graphicx}
   \graphicspath{{../eps/}}
   \DeclareGraphicsExtensions{.eps}
\fi

\usepackage{amsfonts}
\usepackage{amsmath}
\usepackage{subfigure}
\usepackage{steinmetz}
\usepackage{multirow}
\usepackage[table,xcdraw]{xcolor} 
\usepackage{soul}
\usepackage{hyperref}
\soulregister\cite7
\soulregister\eqref7

\begin{document}

\title{FaceRefiner: High-Fidelity Facial Texture Refinement with Differentiable Rendering-based Style Transfer}

\author{Chengyang Li, Baoping Cheng, Yao Cheng, Haocheng Zhang, Renshuai Liu, Yinglin Zheng, Jing Liao, Xuan Cheng
\thanks{

Chengyang Li, Renshuai Liu, Yinglin Zheng and Xuan Cheng are with the School of Informatics, Xiamen University, Xiamen 361005, China (e-mail: chengxuan@xmu.edu.cn).

Baoping Cheng and Yao Cheng are with the China Mobile (Hangzhou) Information Technology Co., Ltd., Hangzhou 311121, China.

Haocheng Zhang is with the School of Computing and Data Science, Xiamen University Malaysia, Sepang 43900, Malaysia.

Jing Liao is with the Department of Computer Science, City University of Hong Kong, Hong Kong 999077, China.

\emph{Corresponding author: Xuan Cheng}.
}
}

\markboth{Journal of \LaTeX\ Class Files,~Vol.~14, No.~8, August~2015}%
{Shell \MakeLowercase{\textit{et al.}}: Bare Demo of IEEEtran.cls for IEEE Journals}

\maketitle

\begin{abstract}
Recent facial texture generation methods prefer to use deep networks to synthesize image content and then fill in the UV map, thus generating a compelling full texture from a single image. Nevertheless, the synthesized texture UV map usually comes from a space constructed by the training data or the 2D face generator, which limits the methods' generalization ability for in-the-wild input images. Consequently, their facial details, structures and identity may not be consistent with the input. In this paper, we address this issue by proposing a style transfer-based facial texture refinement method named FaceRefiner. FaceRefiner treats the 3D sampled texture as \emph{style} and the output of a texture generation method as \emph{content}. The photo-realistic style is then expected to be transferred from the style image to the content image. Different from current style transfer methods that only transfer high and middle level information to the result, our style transfer method integrates differentiable rendering to also transfer low level (or pixel level) information in the visible face regions. The main benefit of such \emph{multi-level} information transfer is that, the details, structures and semantics in the input can thus be well preserved. The extensive experiments on Multi-PIE, CelebA and FFHQ datasets demonstrate that our refinement method can improve the texture quality and the face identity preserving ability, compared with state-of-the-arts. The code is available in \href{https://github.com/HarshWinterBytes/FaceRefiner}
{https://github.com/HarshWinterBytes/FaceRefiner}

\end{abstract}


\section{Introduction}

\IEEEPARstart{H}{igh}-fidelity facial texture generation is an important procedure for human face digitization. Most face photos we take, however, can't exhibit the full view of a face, thus hindering the reconstruction of a complete ear-to-ear texture UV map.
The task of generating facial texture  from an image
requires inferring invisible face content while keeping the full texture image harmonious in the UV space, which has a widely range of applications, from 3D Morphable Model (3DMM) construction \cite{blanz1999morphable,booth20173d,ploumpis2019combining}, 3D avatar creation \cite{ichim2015dynamic,lattas2020avatarme,lin2021meingame} to pose-invariant face recognition \cite{deng2018uv,Hassner_2015_CVPR}.

A couple of deep learning based facial texture generation methods have been proposed in recent years, and gradually become the mainstream in this research field.
The regression-based methods, e.g. UVGAN \cite{deng2018uv} and DSDGAN \cite{DSDGAN2021}, usually gather complete or incomplete texture UV maps as dataset to train a regression network.
The training-free methods, e.g. OSTEC \cite{gecer2021ostec}, step aside from the effort for data collection, and optimize the parameters in a pre-trained 2D face generator, StyleGAN v2 \cite{karras2020analyzing}, to fill in the unseen parts.

%The supervised methods, e.g. UVGAN \cite{deng2018uv},usually gather complete texture UV maps as ground truth and then train an image-to-image translation network. The unsupervised methods, e.g. DSDGAN \cite{DSDGAN2021}, eliminate the need for complete texture acquisition, and employ the differentiable mesh renderer to predict pixels in invisible regions. Both the supervised and unsupervised methods require large-scale facial texture collection and long-time training to establish the regression models. The optimization-based methods, e.g. OSTEC \cite{gecer2021ostec}, step aside from the effort for data collection, and optimize the parameters in a pre-trained 2D face generator, StyleGAN v2 \cite{karras2020analyzing}, to fill in the unseen parts.

\begin{figure*} 
\centering
\includegraphics[width=0.95\textwidth]{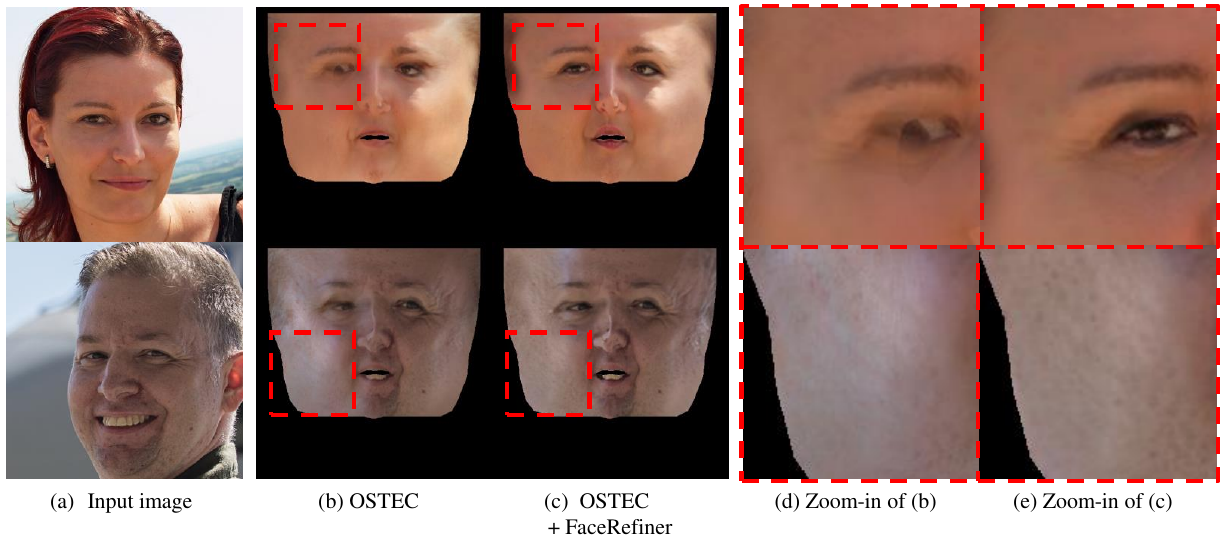} 
\caption{ 
The refined facial textures by our proposed FaceRefiner on the the results produced by OSTEC \cite{gecer2021ostec}, can yield more eye features (1st row) and skin spots (2nd row), and thus better preserve the face identity.}
\label{fig:banner}
\end{figure*}

However, the use of the facial texture dataset or the 2D face generator also limits the generalization ability of the above  methods for processing in-the-wild images. The synthesized texture is actually sampled from a specific data distribution constructed by the training data or StyleGAN v2, to best match the input image. The in-the-wild face images, however, inherently have a much wider domain. Consequently, the details, structures and identity of the synthesized textures may not be consistent with the inputs. We show two examples in Fig. \ref{fig:banner}. The facial textures generated by OSTEC lose the eye features or the skin spots from the input images. This ``\emph{face shifting}'' phenomenon hinders the realism of texture and the identity of the face, especially when the input face is of high quality and has large poses.

To eliminate the face shifting phenomenon existing in the facial texture generation methods, 
we propose a facial texture refinement method named \emph{FaceRefiner} from the perspective of style transfer. FaceRefiner treats the facial texture sampled by 3D face reconstruction as a \emph{style image}, and the output of a facial texture generation method as a \emph{content image}, considering that the 3D sampled texture (incomplete) can preserve more details from the input image than the texture (complete) generated by deep networks. The goal of facial texture refinement becomes to transfer the photo-realistic style from the style image to the content image.
%promote the generalization ability of current facial texture generation methods, 

It is a nontrivial task to adopt style transfer in facial texture refinement. Firstly, as the style image contains large invalid regions in the UV space due to self-occlusion, the style transfer is required to intelligently recognize the invalid regions and transfer information only in the valid regions. Secondly, most style transfer methods \cite{gatys2016image,selim2016painting,li2016combining,champandard2016semantic,johnson2016perceptual,sanakoyeu2018style,kolkin2019style,li2017universal,kalischek2021light, StripsST} can only transfer high level (e.g. colours and lighting in the style image) and middle level (e.g. lines and shapes in the content image) information to the result. When applied in facial texture refinement, however, an ideal style transfer method should also transfer pixel level information, since the fine details and structures are embedded in the raw pixels of input images.

To address these issues, firstly, we have tried out various style transfer methods and choose STROTSS \cite{kolkin2019style} as the backbone, considering that STROTSS can prevent style transferring in invalid regions by using hypercolumn matching. Secondly and importantly, we propose to integrate differentiable rendering in style transfer, through rendering the optimized facial texture back to the input image space, and jointly minimizing the rendering loss, style loss and content loss in a multi-stage framework. In this way, the fine details (e.g. skin details and face spots), the significant structures (e.g. facial hair and eyebrow), and the face identity in the input image can be mostly transferred to the final result.

In summary, the contributions of this paper are as follows.
1) We propose a general and flexible face refinement method, which can act as a post processing operation without training for any facial texture generation method.
2) We re-design the classical style transfer method by incorporating the differentiable rendering to transfer multi-level information from inputs,
thus making it suitable for facial texture migration.
3) We conduct extensive experiments on Multi-PIE, CelebA and FFHQ datasets, which shows significant improvement on texture quality and identity preserving over state-of-the-arts.

%------------------------------------------------------------------------
\section{Related Work}
\label{relatedwork}

\subsection{Facial Texture Generation}
Due to the incredible ability of GAN \cite{karras2020analyzing, karras2017progressive, emef2023}, more and more methods apply GAN for facial texture generation from a single image. UVGAN \cite{deng2018uv} and GANFIT \cite{gecer2019ganfit} can generate facial texture of rich details, but require a large amount of complete texture data for training and thus are limited by the parametric model space. DSDGAN \cite{DSDGAN2021} does not need complete texture acquisition, but still requires large-scale data collection (natural facial images) and long-time training. To get rid of the dependence on training data, OSTEC \cite{gecer2021ostec} adopts a one-shot optimization framework. Profit from the power of StyleGAN v2 \cite{karras2020analyzing}, their method can infer high-resolution and detailed results. Similarly, MvInvert \cite{chen2022towards} leverages the residual-based latent encoder \cite{alaluf2021restyle} and StyleGAN v2 to obtain good textures of multiple views, and then fuses them to get the high-fidelity textures. Noteworthy, because of the data space in StyleGAN v2, there may be a loss of identity information. The recent methods DCT \cite{DCT2023} and DNPM \cite{DNPM} use Transformer and StyleGAN to generate a special type of texture, not color image but the displacement map.
Instead of generating the textures from scratch, the proposed FaceRefiner 
acts as a post processing operation to refine the generated textures, after unifying the UV coordinates.

\begin{figure*} 
\centering
\includegraphics[width=0.98\textwidth]{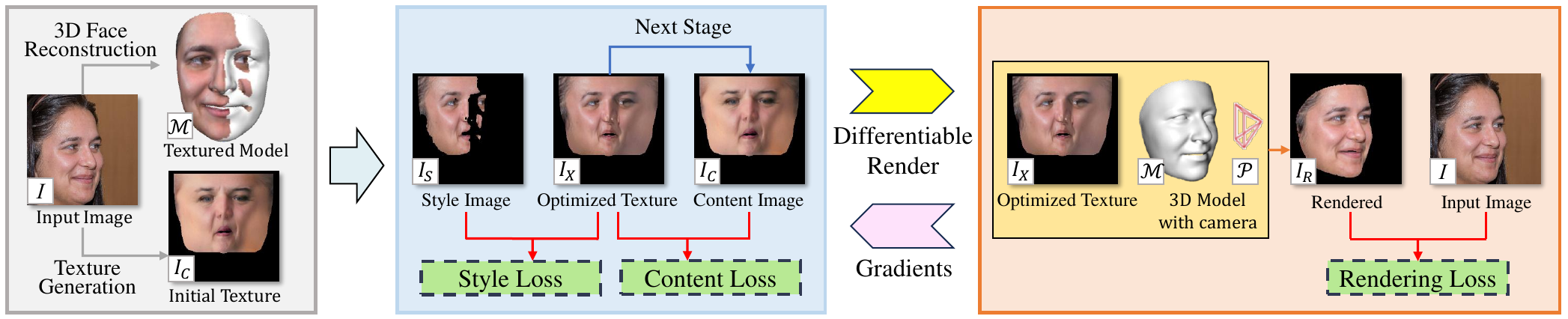} 
\caption{ 
The overview of our proposed FaceRefiner. The inputs of FaceRefiner include the face image $I$, the 3D face reconstruction results (3D model $\mathcal{M}$ and camera pose $\mathcal{P}$, sampled texture $I_S$) and the initial imperfect texture $I_C$ produced by an existing facial texture generation method. The differentiable rendering-based style transfer is adopted to improve the quality of $I_C$. The differentiable renderer is employed to produce rendered image $I_R$ of the inputted camera pose $\mathcal{P}$. Then the rendering loss is calculated to measure the inconsistency between rendered and inputted image, and the gradients are back-propagated to a classical style transfer module containing style and content loss to optimize the facial texture $I_X$.}
\label{fig:overview}
\end{figure*}

\subsection{Style Transfer}
Style transfer aims to render the content of one image using the style of another. Although style transfer algorithms \cite{hertzmann2001image,shih2013data} have existed for decades, it was not until 2016 Gatys et al. introduced Neural Style Transfers \cite{gatys2016image}. In the same year, a large number of works \cite{selim2016painting,li2016combining,champandard2016semantic} that improve upon \cite{gatys2016image} emerged. Subsequently, researchers have continued to improve the methods of their predecessors and explore new paths from different perspectives. For example, because optimization-based methods are computationally intensive, some faster regression models \cite{johnson2016perceptual,sanakoyeu2018style} have been proposed. 
There are also methods proposed to tackle the limitation of pre-defined styles \cite{li2017universal}, match the feature distributions more precisely \cite{kalischek2021light}, represent features not with gram matrix but with hypercolumn \cite{kolkin2019style}, process the style image in a sequential strips way \cite{StripsST}, separate the input into texture and structure \cite{TexturePreservingST2022}, and explore the temporal consistency in style transfer \cite{kong2023exploring}. Others find that previous work had mostly limited style to textures and colors, so they expand the definition of style to enable the migration of shapes \cite{kim2020deformable}, lines \cite{liu2021deep}, etc. In recent years, the vehicles of style and content have begun to be replaced by forms other than images, such as using text and fonts to represent style \cite{li2020fet}, or migrating style to video \cite{deng2021arbitrary, VideoST2021}. In this paper, we adopt the basic image-to-image style conversion, and choose STROTSS \cite{kolkin2019style} as the backbone style transfer method, since STROTSS uses hypercolumn matching to achieve the style transfer only in the valid regions.

\subsection{Image Inpainting}
Image inpainting method can also be used to generate facial texture from an incomplete input. Many inpainting methods based on deep neural networks have emerged in the past few years. Yu et al. \cite{yu2018generative} proposed an end-to-end image inpainting model by adopting stacked generative networks and a contextual attention module. However, it can only fill rectangular holes. 
In the same year, Liu et al. \cite{liu2018image} proposed partial convolution where the convolution is masked and consider only valid pixels to operate robustly on free-form holes. To improve image inpainting, a generative image inpainting method based on gated convolutions was proposed by Yu et al. \cite{yu2019free} to deal with irregular masks and guidance. Later, noting that image structure knowledge is not well explored, Yang et al \cite{yang2020learning} trained a shared generator to exploit relevant structure knowledge to assist inpainting. When applied in facial texture generation, the image inpainting methods usually can't achieve global smoothness in the UV space.

\section{Method}
\subsection{Overview}
\textbf{Problem Setting.} As shown in the gray box in Fig. \ref{fig:overview}, given an input face image $I$, the 3D face reconstruction is conducted to obtain the 3D face model $\mathcal{M}$ with incomplete texture $I_S$. Meanwhile, the facial texture generation method like \cite{gecer2021ostec, deng2019accurate} is used to produce the complete facial texture $I_C$.
The goal of our refinement method is to improve the quality of $I_C$, only based on $I$, $\mathcal{M}$ and $I_S$. 

\begin{figure} 
\centering
\includegraphics[width=0.48\textwidth]{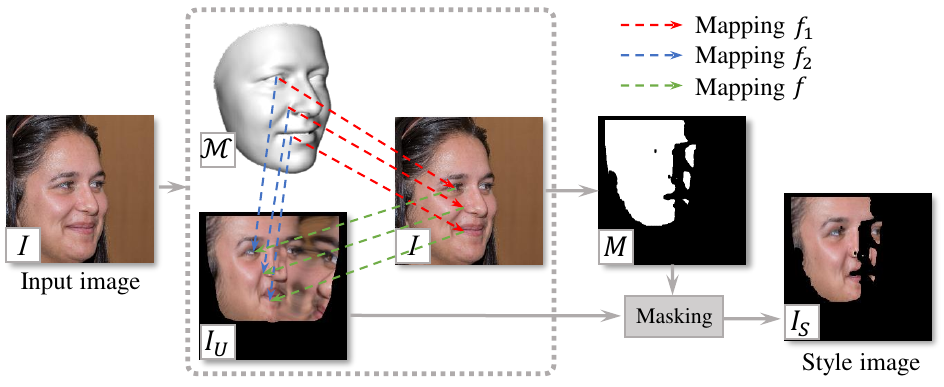} 
\caption{The generation of style image.}
\label{fig:styleimage}
\end{figure}

\textbf{Pipeline.} To realize the goal, we propose the differentiable rendering-based style transfer, which is shown in the blue and orange boxes in Fig. \ref{fig:overview}. Following the standard setting of style transfer, $I_S$ serves as the style image and $I_C$ serves as the content image. Then, the texture $I_X$ is optimized by minimizing the style loss and content loss. 
Through the differentiable renderer, $I_X$ is rendered into the input view in the light of the estimated 3D face model $\mathcal{M}$ and camera pose $\mathcal{P}$. The rendering loss measuring the inconsistency between rendered image $I_R$ and input image $I$ can thus be calculated, and back-propagated by the differentiable renderer to update the texture $I_X$. Such style transfer is conducted in a multi-stage manner, by treating $I_X$ in the current stage as the content image in the next stage.

\subsection{Style and Content Image Generation}
\label{sect:styleimage}
\textbf{3D Face Reconstruction.} We utilize the state-of-the-art 3D face reconstruction method \cite{deng2019accurate} to estimate 3D face model $\mathcal{M}$ and camera pose 
$\mathcal{P}$ from the input image $I$. Then, based on the face reconstruction results, three mappings $f_1, f_2, f$ are calculated. 
As shown in Fig. \ref{fig:styleimage}, $f_1$ represents the mapping from each vertex in $\mathcal{M}$ to its corresponding pixel in $I$, $f_2$ represents the mapping from each vertex in $\mathcal{M}$ to its corresponding pixel in the sampled UV texture $I_U$.
Regarding $\mathcal{M}$ as a ``bridge'', we can obtain the mapping $f$ through the combination of $f_{1}$ and $f_{2}$.

\textbf{Style Image.} Due to the face self-occlusions in $I$, the invisible regions in $I_U$ may be incorrectly sampled from $I$, which results in a large number of distorted and anamorphosis regions in $I_{U}$. To eliminate these errors, we compute the visibility of each vertex of $\mathcal{M}$ and obtain the visibility mask $M$. 
Firstly, we compute the dot product of the camera view and the facet normal and empirically consider the vertices with a value less than 0.6 as invisible, otherwise visible. Then, to smooth the boundary and fill small holes in the visibility mask, we conduct a series of morphological opening and closing operations. In this way, the style image $I_S$ can be defined as:
\begin{equation}
    I_S = f(I) \odot O(M')
\end{equation}
where $O(\cdot)$ denotes the set of the morphological opening and closing operations, $\odot$ denotes the Hadamard product, and $M'$ denotes the initial mask with rough boundary and small holes. The invalid pixels in $I_S$ are denoted in black color.

\textbf{Content Image.} As reviewed in Sect. \ref{relatedwork}, a couple of deep learning-based facial generation methods can be used to complete $I_S$, thus producing the content image $I_C$. 
Although the completion operation can synthesize image content in invisible regions, it also causes information loss in several aspects, making $I_C$ not ``look like'' the input $I$.
% $I_C$ also contains invalid pixels, which are masked by black color. 

\subsection{Differentiable Rendering-based Style Transfer}
\label{texturemigration}
\textbf{Backbone.}
We choose the style transfer method STROTSS \cite{kolkin2019style} as the backbone. Although there exists many other style transfer methods, such as WCT \cite{li2017universal}, CMD \cite{kalischek2021light} and StyTR2 \cite{deng2022stytr2}, only STROTSS is effective in transferring information from incomplete style image. More analyses and comparisons of different style transfer methods can be found in Sect. \ref{sect:discussion} and \ref{sect:compareStyleTransfer}. The hypercolumn \cite{hariharan2015hypercolumns, mostajabi2015feedforward} is used to extract the image features, since it contains multiple levels of semantics.

\textbf{Overall Loss.} 
The loss function $L_{st}$ in style transfer is defined as:
\begin{equation}
\label{E5}
    L_{st} = \alpha L_{content} + \beta L_{style} + \gamma L_{render},
\end{equation}
where $\alpha$, $\beta$ and $\gamma$ are the hyperparameters of three loss terms. 
The classical content loss $L_{content}$ and style loss $L_{style}$ together transfer the high level and middle level information from $I_S$ and $I_C$ to result $I_X$, while the rendering loss $L_{render}$ complementarily transfers the pixel level information from $I$ to $I_X$.

\textbf{Content Loss.} 
Based on the self-similarity, $L_{content}$ is defined as the absolute error between the cosine distances of the normalized pairs of feature vectors extracted from the content image $I_C$ and the output image $I_X$ respectively. It can be formulated by:
\begin{equation}
\label{E6}
    L_{content}(I_X, I_C) = \frac{1}{n^{2}} \sum_{i,j}\left| D_{i,j}^{X} - D_{i,j}^{C}\right|,
\end{equation}
where $D^{X}$ is the cosine distance matrix of paired feature vectors of the $I_X$, $D^{C}$ is similarly defined to $I_C$, and $n$ is the number of rows and columns of the feature matrix.
$L_{content}$ ensures that the output image and the content image are similar in spatial structure.

\textbf{Style Loss.} 
$L_{style}$ includes three terms and is defined as:
\begin{equation}
\label{E7}
    L_{style}(I_X, I_S) = L_{r} + L_{m} + \frac{1}{max(\alpha, 1)}L_{p}.
\end{equation}
$L_{r}$ is the Earth Mover's Distance (EMD) between $I_X$ and $I_S$,  which is adept in migrating the texture from $I_S$ to $I_X$. As EMD is actually the cosine distance, the relationship between the feature vector lengths is ignored, causing artifacts in the result. The moment matching loss $L_{m}$ \cite{kolkin2019style} is adopted to address this issue. The color matching loss $L_{p}$ \cite{kolkin2019style} ensures that $I_X$ is as similar as possible to $I_S$ in terms of color. In particular, $\alpha$ is the same as the hyperparameter of $L_{content}$ in Eq. \ref{E5}.

\textbf{Rendering Loss.} 
Neither $L_{content}$ nor $L_{style}$ constrains the $I_X$ at the pixel level, and thus inevitably leads to significant visual difference between $I_X$ and $I$. To compute the rendering loss 
$L_{render}$ in the pixel level, the 3D face model $\mathcal{M}$, the camera pose $\mathcal{P}$ and the optimized texture $I_X$ are fed to the differentiable renderer $DR$ to generate the color image $I_R$. The rendering operation can formulated by:
\begin{equation}
\label{eq:rendering}
   I_R = DR(\mathcal{M}, I_X | \mathcal{P}).
\end{equation}
$DR$ firstly projects the vertices in $\mathcal{M}$ into the image space in the light of the camera pose $\mathcal{P}$, then adds the spatially-varying factors, such as texture map $I_X$ and lighting model, on the pixels in image space, finally conducts the 2D antialiasing filter on the shading result. The gradients can be back-propagated in $DR$ during optimization. More implementation details about differentiable renderer can be found in the literature \cite{laine2020modular}.

$L_{render}$ is defined by measuring the difference between $I_R$ and its ground truth $I$:
\begin{equation}
\begin{split}
L_{render}(I_X, I) = \sum_{i}
\left|(I_{R, i} - I_{i}) \odot M_{F_{i}}\right|,
\end{split}
\end{equation}
where $i$ denotes the pixel index and $M_F$ denotes the face area in the $I$. 

Particularly, to transfer pixel level information, we do not directly calculate the reconstruction loss between $I_X$ and $I_S$ in the UV space. Through experiments we find that, $I_S$ contains fewer valid face pixels around the face boundary than $I$, due to the series of morphological operations conducted in the visibility mask $M'$. In the ablation study, we will show the advantage of the rendering loss over the UV reconstruction loss.

\textbf{Multi-stage Transfer.}
To further enforce the pixel consistency between $I_X$ and $I$, we conduct multi-stages of style transfer. In each stage $i$, the input content image $I_{C}^i$ is the output in the previous stage $I_X^{i-1}$, while the style image $I_S$ is kept fixed throughout. 
In order to speed up the convergence, we take the Laplacian pyramid of the content image $I_C$ as the initial parameters to be optimized instead of pixels, following the previous work \cite{kolkin2019style}. 
Every stages of style transfer are roughly the same except for the hyperparameters updating: 
%(1) the style image is flipped horizontally at the beginning of the even-numbered migrations to fully migrate the texture information; (2)
$\alpha$ and $\gamma$ are adjusted as $\alpha_{i + 1} = \alpha_{i} / 1.1, \gamma_{i + 1} = \gamma_{i} * 1.1$ at the end of each stage.

%\hl{\textbf{Differentiable Renderer.}}
%A differentiable renderer is a technique that can quickly render the scene in 3D space into image space and ensure the entire process is differentiable which means it can be used to optimize the neural network by backpropagation techniques. The rendering process can be broken down into the following form:
%\begin{equation}
%\label{eq:rendering}
%   C_{x, y} = AF(S(SF(P(v)), l)).
%\end{equation}
%where $C_{x, y}$ denotes the final color of screen coordinate $(x, y)$. $P(\cdot)$ project the vertex in 3D space into image space and $SF(\cdot)$ add spatially-varying factors, such as texture maps and normal vectors, on the surface of the scene. $S(\cdot, l)$ denotes the shade operation that models light-surface interactions and $l$ denotes the light parameters. Finally, the 2D antialiasing filter $AF(\cdot)$  is employed to the shading results. For more details, please refer to \cite{laine2020modular}

\subsection{Discussion and Analysis.}
\label{sect:discussion}
\begin{figure} 
\centering
\includegraphics[width=0.46\textwidth]{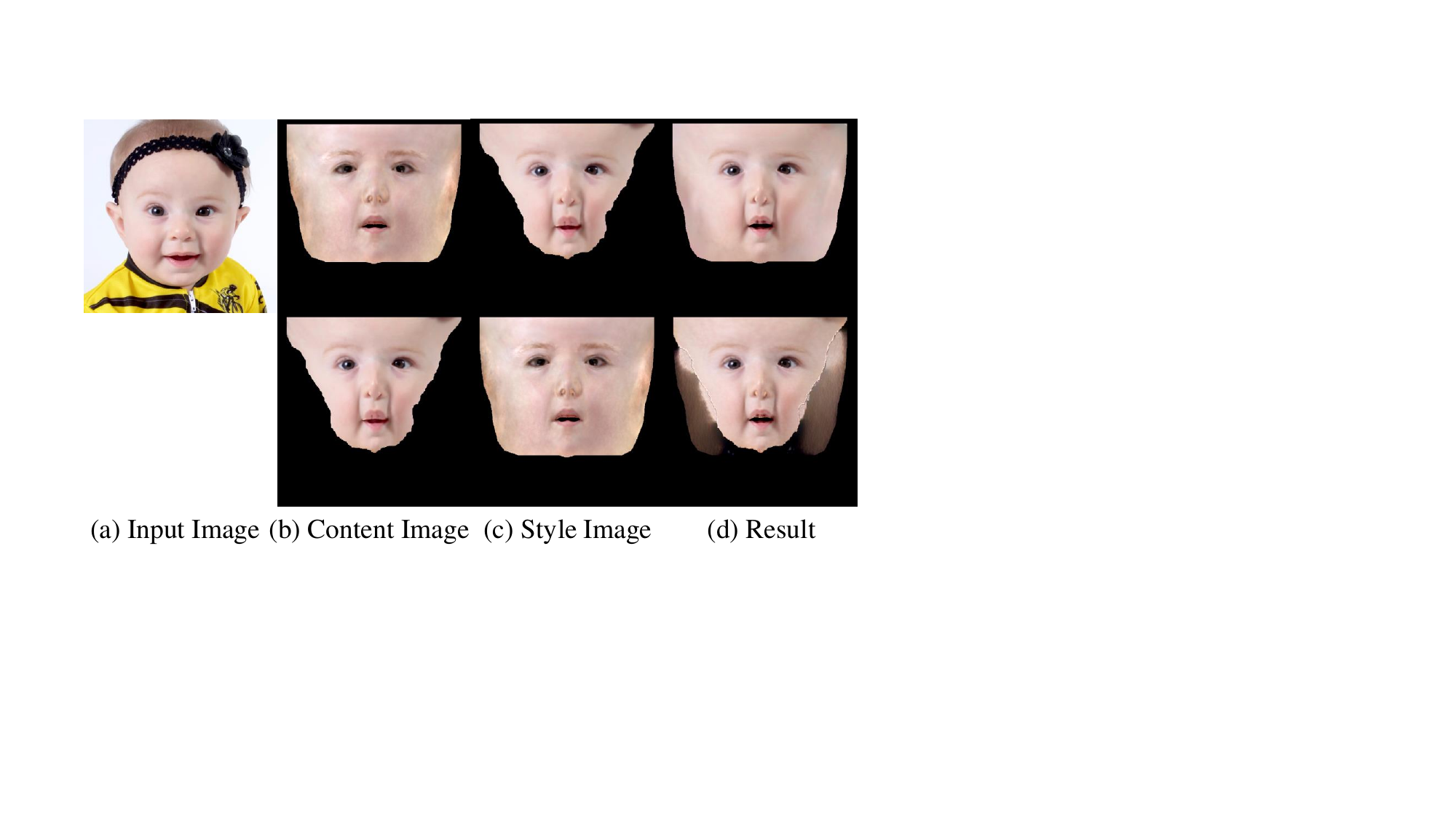}
\caption{In 1st row, the style and content images are generated in the proposed way. In 2nd row, the style and content images are swapped.}
\label{fig:swapping}
\end{figure}

\textbf{Swapping style and content image.} The style and content images which are generated in the proposed way (Sect. \ref{sect:styleimage}) work well in the texture refinement task. As shown in Fig. \ref{fig:swapping}, if we swap the style and content images, the results are unreasonable. The style transfer does not have the ability of synthesizing pixels in the missing regions.

\begin{figure} 
\centering
\includegraphics[width=0.46\textwidth]{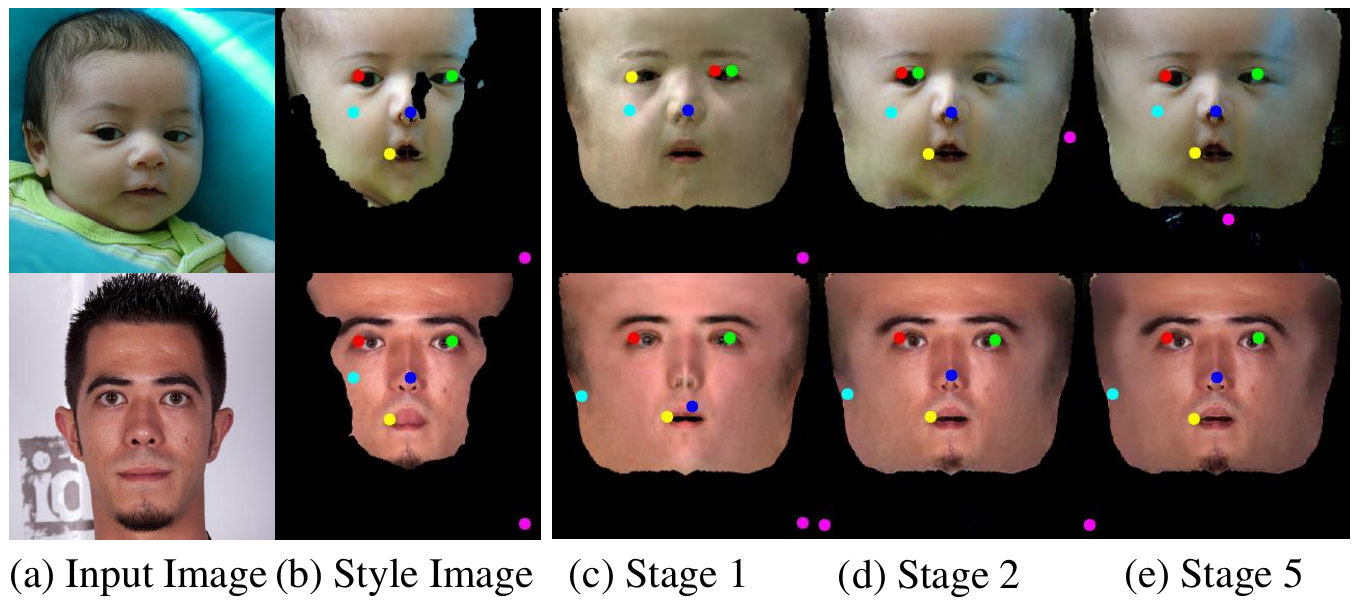} 
\caption{The illustration of hypercolumn matching between style image and output image. (c)-(e) show the output images produced by performing different stages of optimization.}
\label{fig:hypercolumn}
\end{figure}

\textbf{Why do we choose STROTSS?} The main advantage of STROTSS over other methods is that, it supports ROI (Region of Interest) style transfer \cite{kolkin2019style}. For each sample point $p_{i}$ in the output image $I_X$, STROTSS selects the point in the style image $I_S$ that has the most similar hypercolumn with $p_{i}$ to compute the style loss. In this way, the invalid pixels (denoted in black color) in $I_S$ will not be transferred and corrupt the valid regions in $I_X$. We conduct a simple experiment to support this claim. As shown in Fig. \ref{fig:hypercolumn}, we select six representative points in $I_S$, including left eye, right eye, mouth corner, nose tip, cheek and an invalid point, and then use hypercolumn matching to find their corresponding point (denoted in the same color) in different $I_X$ produced by performing different stages of optimization. At the beginning of optimization, these matchings are probably wrong. With the going of multi-stage optimization, they become progressively more correct. The invalid point (pink color) in $I_S$ will not correspond to any point in the valid regions of $I_X$.

\begin{figure} 
\centering
\includegraphics[width=0.38\textwidth]{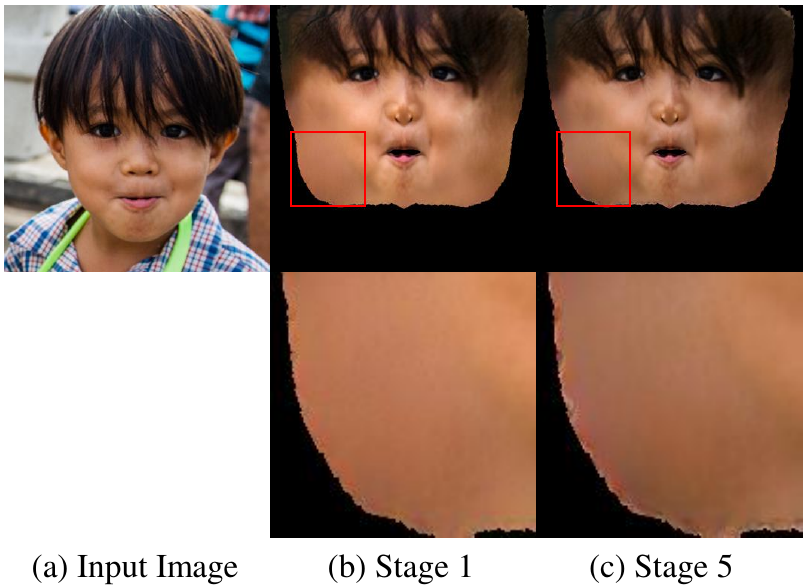} 
\caption{In our method, the content leak of small scale appears in the texture boundary. (b)(c) show the refined textures by using our method with one-stage and five-stage optimization.}
\label{fig:content_leak}
\end{figure}

\textbf{Content Leak.} The content leak issue is about the image content corruption in the output image $I_X$ caused by performing multiple stages of stylization, which is well discussed in the literature \cite{Artflow2021, deng2022stytr2}. As shown in Fig. \ref{fig:content_leak} (b)(c), the content leak phenomenon of small scale is observed in our method, where the boundary of the refined texture is distorted after five stages of stylization. To alleviate the content leak, we take a intuitive method which uses a defaulting mask on the refined texture to remove the distorted pixels around boundary.

%As StyTR2 \cite{deng2022stytr2} can address the content leak issue by using Transformer, we make a attempt to adopt it in our task. As shown in Fig. \ref{fig:content_leak} (d)(e), the distortion in the boundary still exists. 

\section{Experiments}

\label{S4}
\subsection{Implementation Details}
We select a total of 2179 feature maps output from 9 sub-layers of VGG16 trained on ImageNet \cite{simonyan2014very} to capture the hypercolumns and represent the image features.
We set the initial values of the hyperparameters as $\alpha = 8.0$, $\beta = 1.7$, and $\gamma = 20.0$ respectively. The number of stages in style transfer is experimentally set as 5.
In each stage, (1) we use the stochastic gradient descent (SGD) optimizer, with 150 iterations, the learning rate of 0.3, and the momentum of 0.9; 
(2) we incrementally work on two scales: 256 × 256 and 512 × 512, and make $\alpha$, $\beta$, $\gamma$ in Eq. \ref{E5} decrease dynamically with the increasing of the scale. 
We implement our method with PyTorch in the NVIDIA RTX 3090 GPU. 
With each stage taking about 24 seconds to refine a $512 \times 512$ facial texture, the total running time of our FaceRefiner is about 2 minutes.

\subsection{Experimental Settings}
\textbf{Evaluation Datasets and Metrics.} In the quantitative experiments, we evaluate all methods in two datasets, Multi-PIE \cite{gross2010multi} and CelebA \cite{liuICCV2015}. 

1) Multi-PIE dataset contains multi-view face images and their corresponding complete facial UV texture. From Multi-PIE, 100 faces of each pose from [-60°, -30°, 0°, +30°, +60°] are selected as the input images. Totally, the testing dataset includes 500 images. PSNR (Peak Signal to Noise Ratio) \cite{psnr2010} and SSIM (Structural Similarity) \cite{wang2004image} are utilized as the evaluation metrics, which are calculated between the inferred face UV texture and the ground truth. The higher PSNR and SSIM, the better generation performance.

\begin{table*}
\centering
\begin{tabular}{c|cccc|cccc}
\hline
\multicolumn{1}{l|}{}          & \multicolumn{4}{c|}{PSNR $\uparrow$}                                                                                                                                   & \multicolumn{4}{c}{SSIM $\uparrow$}                                                                                                         \\ \hline
                               & 0°                              & ±30°                            & ±60°                                                   & mean                           & 0°                             & ±30°                           & ±60°                           & mean                          \\ \hline
Deep3DFace                     & 17.7575                         & 18.9764                         & 17.7531                                                & 18.2433                        & 0.8295                         & 0.8355                         & 0.8141                         & 0.8257                        \\
OSTEC                          & 25.5208                         & 23.3888                         & 19.8525                                                & 22.4007                        & 0.8763                         & 0.8561                         & 0.8036                         & 0.8391                        \\
\cellcolor[HTML]{FFFFFF}PICNet & \cellcolor[HTML]{FFFFFF}25.6186 & \cellcolor[HTML]{FFFFFF}24.6299 & \cellcolor[HTML]{FFFFFF}{\color[HTML]{3531FF} 22.7184} & 24.0631                        & \cellcolor[HTML]{FFFFFF}0.8645 & \cellcolor[HTML]{FFFFFF}0.8539 & \cellcolor[HTML]{FFFFFF}0.8449 & 0.8524                        \\
Deepfill\_v2                   & 24.8568                         & 23.5752                         & 22.3100                                                & 23.3255                        & 0.8597                         & 0.8464                         & 0.8425                         & 0.8475                        \\ \hline
Deep3DFace + WCT               & \cellcolor[HTML]{FFFFFF}16.2837 & \cellcolor[HTML]{FFFFFF}14.3381 & \cellcolor[HTML]{FFFFFF}14.0064                        & 14.5946                        & \cellcolor[HTML]{FFFFFF}0.7055 & \cellcolor[HTML]{FFFFFF}0.6384 & \cellcolor[HTML]{FFFFFF}0.6148 & 0.6424                        \\
OSTEC + WCT                    & 17.0068                         & 14.7301                         & 14.2572                                                & 14.9963                        & 0.7157                         & 0.6439                         & 0.6115                         & 0.6453                        \\
Deep3DFace + CMD               & \cellcolor[HTML]{FFFFFF}20.7723 & \cellcolor[HTML]{FFFFFF}20.6618 & \cellcolor[HTML]{FFFFFF}19.5827                        & 20.2522                        & \cellcolor[HTML]{FFFFFF}0.8312 & \cellcolor[HTML]{FFFFFF}0.8188 & \cellcolor[HTML]{FFFFFF}0.8015 & 0.8144                        \\
OSTEC + CMD                    & {\color[HTML]{000000} 23.4691}  & {\color[HTML]{000000} 22.3046}  & {\color[HTML]{000000} 18.9365}                         & {\color[HTML]{000000} 21.1903} & {\color[HTML]{000000} 0.8508}  & {\color[HTML]{000000} 0.8238}  & {\color[HTML]{000000} 0.7604}  & {\color[HTML]{000000} 0.8038} \\
Deep3DFace + StyTR2            & 22.8756                         & 19.1120                         & 17.5275                                                & 19.2309                        & 0.8436                         & 0.7853                         & 0.7603                         & 0.7870                        \\
OSTEC + StyTR2                 & 24.7362                         & 20.7043                         & 17.6308                                                & 20.2813                        & 0.8702                         & 0.8111                         & 0.7455                         & 0.7967                        \\
%Deep3DFace + Doodle            & 22.9747                         & 21.3904                         & 19.6422                                                & 21.0080                        & 0.7999                         & 0.7828                         & 0.7455                         & 0.7713                        \\
%OSTEC + Doodle                 & 24.3130                         & 21.6001                         & 19.6165                                                & 21.3493                        & 0.8280                         & 0.7938                         & 0.7538                         & 0.7846                        \\
Deep3DFace + STROTSS           & {\color[HTML]{000000} 17.4634}  & {\color[HTML]{000000} 18.6251}  & {\color[HTML]{000000} 17.1229}                         & {\color[HTML]{000000} 17.7919} & {\color[HTML]{000000} 0.8306}  & {\color[HTML]{000000} 0.8380}  & {\color[HTML]{000000} 0.8157}  & {\color[HTML]{000000} 0.8276} \\
OSTEC + STROTSS                & 20.0058                         & 20.2149                         & 18.4757                                                & 19.4774                        & 0.8531                         & 0.8444                         & 0.8032                         & 0.8296                        \\ \hline
Deep3DFace + FaceRefiner       & {\color[HTML]{FF0000} 27.0525}  & {\color[HTML]{FF0000} 25.8852}  & 22.5371                                                & {\color[HTML]{3531FF} 24.7794} & {\color[HTML]{3531FF} 0.8879}  & {\color[HTML]{FF0000} 0.8778}  & {\color[HTML]{FF0000} 0.8631}  & {\color[HTML]{FF0000} 0.8739} \\
OSTEC + FaceRefiner            & {\color[HTML]{3531FF} 27.0096}  & {\color[HTML]{3531FF} 25.5164}  & {\color[HTML]{FF0000} 22.9465}                         & {\color[HTML]{FF0000} 24.7871} & {\color[HTML]{FE0000} 0.8879}  & {\color[HTML]{3531FF} 0.8753}  & {\color[HTML]{3531FF} 0.8600}  & {\color[HTML]{3531FF} 0.8717} \\ \hline
\end{tabular}
\vspace{2mm}
\caption{
The quantitative results of the facial texture generation methods, image inpainting methods and other style transfer methods over the Multi-PIE dataset. 
The best score in each column is colored with {\color[HTML]{FE0000} red}, and the second-best is colored with {\color[HTML]{3531FF} blue}.}
\label{tab:MultiPIE}
\end{table*}

\begin{table*}
\centering
\vspace{-4mm}
\begin{tabular}{c|cc|c|cc}
\hline
                             & \multicolumn{2}{c|}{Pixel-Level}                               & Middle-Level                                      & \multicolumn{2}{c}{High-Level}                           \\ \hline
& MAE $\downarrow$                           
& PSNR $\uparrow$                           
& SSIM $\uparrow$                                         
& LightCNN $\uparrow$                      
& evoLVe $\uparrow$                         \\ \hline
Deep3DFace                   & 0.0325                         & 23.6582                         & 0.8346                                                & 0.6896                        & 0.6271                         \\
OSTEC                        & 0.0256                         & 25.6900                         & 0.8841                                                & 0.8319                        & 0.7982                         \\
\cellcolor[HTML]{FFFFFF}3DFaceGCNs & \cellcolor[HTML]{FFFFFF}0.0340 & \cellcolor[HTML]{FFFFFF}29.6900 & \cellcolor[HTML]{FFFFFF}{\color[HTML]{000000} 0.8940} & 0.9000                        & \cellcolor[HTML]{FFFFFF}0.8480 \\
MvInvert                     & 0.0280                         & {\color[HTML]{FF0000} 30.7800}  & 0.8970                                                & 0.9260                        & 0.8780                         \\ \hline
\rowcolor[HTML]{FFFFFF} 
PICNet                       & 0.0163                         & {\color[HTML]{000000} 27.7602}  & 0.9162                                                & 0.9308                        & 0.9022                         \\
\rowcolor[HTML]{FFFFFF} 
Deepfill\_v2                 & 0.0170                         & 27.1759                         & 0.9118                                                & 0.9369                        & 0.9140                         \\ \hline
Deep3DFace + WCT             & \cellcolor[HTML]{FFFFFF}0.0896 & \cellcolor[HTML]{FFFFFF}15.0115 & \cellcolor[HTML]{FFFFFF}0.6713                        & 0.2650                        & \cellcolor[HTML]{FFFFFF}0.1466 \\
OSTEC + WCT                  & {\color[HTML]{000000} 0.0805}  & {\color[HTML]{000000} 15.9290}  & {\color[HTML]{000000} 0.6960}                         & {\color[HTML]{000000} 0.3293} & {\color[HTML]{000000} 0.1990}  \\
Deep3DFace + CMD             & {\color[HTML]{000000} 0.0413}  & {\color[HTML]{000000} 21.8956}  & {\color[HTML]{000000} 0.8074}                         & {\color[HTML]{000000} 0.5979} & {\color[HTML]{000000} 0.5008}  \\
OSTEC + CMD                  & 0.0438                         & 21.9862                         & 0.8328                                                & 0.7111                        & 0.6329                         \\
Deep3DFace + StyTR2          & 0.0418                         & 21.7763                         & 0.8042                                                & 0.6175                        & 0.5540                         \\
OSTEC + StyTR2               & 0.0440                         & 21.8140                         & 0.8434                                                & 0.7610                        & 0.7216                         \\
%Deep3DFace + Doodle          & 0.0342                         & 23.1986                         & 0.8212                                                & 0.6705                        & 0.6056                         \\
%OSTEC + Doodle               & 0.0318                         & 24.0646                         & 0.8602                                                & 0.7996                        & 0.7536                         \\
Deep3DFace + STROTSS         & {\color[HTML]{000000} 0.0350}  & {\color[HTML]{000000} 23.3212}  & {\color[HTML]{000000} 0.8359}                         & {\color[HTML]{000000} 0.6996} & {\color[HTML]{000000} 0.6451}  \\
OSTEC + STROTSS              & {\color[HTML]{000000} 0.0314}  & {\color[HTML]{000000} 24.4375}  & {\color[HTML]{000000} 0.8820}                         & {\color[HTML]{000000} 0.8360} & {\color[HTML]{000000} 0.7991}  \\ \hline
Deep3DFace + FaceRefiner     & {\color[HTML]{3531FF} 0.0142}  & 30.1138                         & {\color[HTML]{3531FF} 0.9344}                         & {\color[HTML]{3531FF} 0.9812} & {\color[HTML]{3531FF} 0.9784}  \\
OSTEC + FaceRefiner          & {\color[HTML]{FF0000} 0.0140}  & {\color[HTML]{3531FF} 30.1968}  & {\color[HTML]{FF0000} 0.9375}                         & {\color[HTML]{FF0000} 0.9853} & {\color[HTML]{FF0000} 0.9828}  \\ \hline
\end{tabular}
\vspace{2mm}
\caption{
The quantitative results of the facial texture generation methods, image inpainting methods and other style transfer methods over the CelebA dataset.}
\label{tab:CelebA}
\end{table*}

2) CelebA dataset contains massive celebrity images, and 500 images are randomly selected from CelebA as the testing dataset. The textures generated by the evaluated methods are projected back into the input image space and then contrasted to the ground truth in pixel-level, middle-level and high-level. The pixel-level evaluation metrics include MAE (Mean Absolute Error) and PSNR, the middle-level evaluation metrics include SSIM, and the high-level (face identity) evaluation metrics are defined by calculating the cosine similarity in the features extracted respectively by two face recognition networks, LightCNN \cite{wu2018light} and evoLVe \cite{zhao2019multi}. The higher scores in LightCNN and eoLVe metrics, the better face identity preserving. The above evaluation metrics can well measure the performance in multi-level information transfer.

\textbf{Competitors.} We choose three types of methods as the competitors, including:
1) facial texture generation methods, which generate facial texture $I_X$ from the input image $I$; 2) image inpainting methods, which generate facial texture $I_X$ from the incomplete texture $I_S$; 3) style transfer methods, which generate facial texture $I_X$ by conducting style transfer on the content image $I_C$ and the style image $I_S$.

\subsection{Comparison with facial texture generation methods}
We select several recently published facial texture generation methods as the competitors to make the quantitative evaluation, including Deep3DFace \cite{deng2019accurate}, OSTEC \cite{gecer2021ostec}, 
3DFaceGCNs \cite{lin2020towards} and MvInvert \cite{chen2022towards}.

The code of Deep3DFace and OSTEC are published online, and we use their code for testing on both Multi-PIE and CelebA. The public code of 3DFaceGCNs and MvInvert can not be directly used, due to the lack of some necessary pre-trained models and usage instructions. We use the quantitative evaluation results in CelebA reported in their paper. Other methods, such as UVGAN \cite{deng2018uv}, DSDGAN \cite{DSDGAN2021} and GANFit \cite{gecer2019ganfit}, do not publicly provide their code, pre-trained models or testing data, making quantitative comparison impossible. For these three methods, we only intercept some of the results in their papers for qualitative comparisons. 

The quantitative evaluation results on Multi-PIE and CelebA are presented in Table \ref{tab:MultiPIE} and \ref{tab:CelebA} respectively. In Table \ref{tab:MultiPIE}, our face refinements on Deep3DFace and OSTEC improve the PSNR and SSIM significantly in all face poses. In Table \ref{tab:CelebA}, our face refinements on Deep3DFace and OSTEC bring improvements in pixel-level, middle-level and high-level metrics. It's worthy noting that, our method outperforms all the facial texture generation methods in the identity metrics by a very large margin. 

%not only in the pixel metrics but also in the identity metrics greatly. Besides, the combination of our FaceRefiner with Deep3DFace or OSTEC outperforms 3DFaceGCNs and MvInvert in identity metrics by a large margin.

\begin{figure*}
\vspace{-4mm}
\centering
\includegraphics[width=0.8\textwidth]{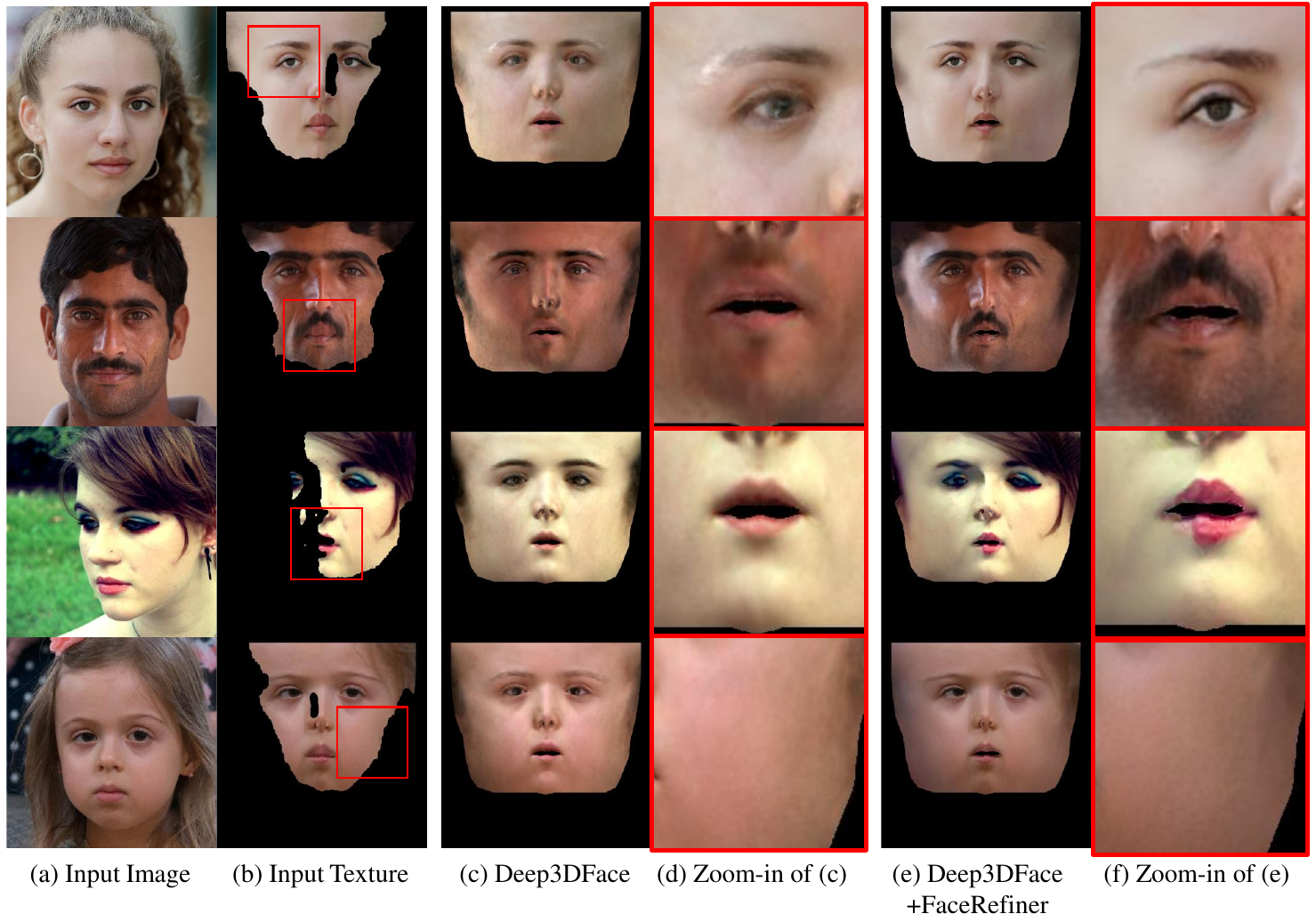} 
\caption{Qualitative comparison between Deep3DFace and Deep3DFace+FaceRefiner on several images from FFHQ.}
\label{fig:coarse_content}
\end{figure*}

\begin{figure*}
\centering
\includegraphics[width=0.8\textwidth]{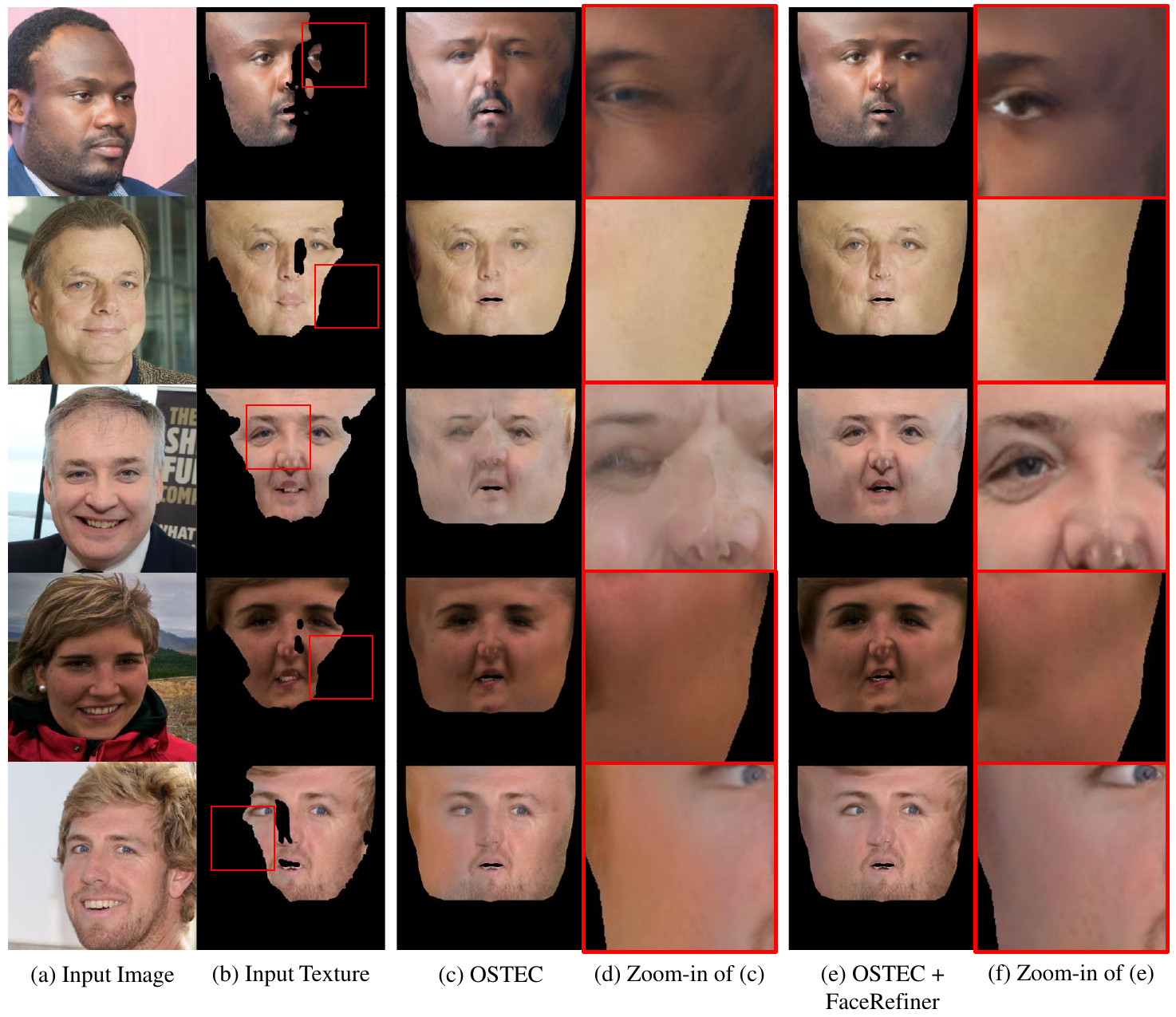} 
\caption{Qualitative comparison between OSTEC and OSTEC+FaceRefiner on several images from FFHQ.}
\label{fig:FFHQ_qualitative}
\end{figure*}

\begin{figure*}
\centering
\includegraphics[width=1.0\textwidth]{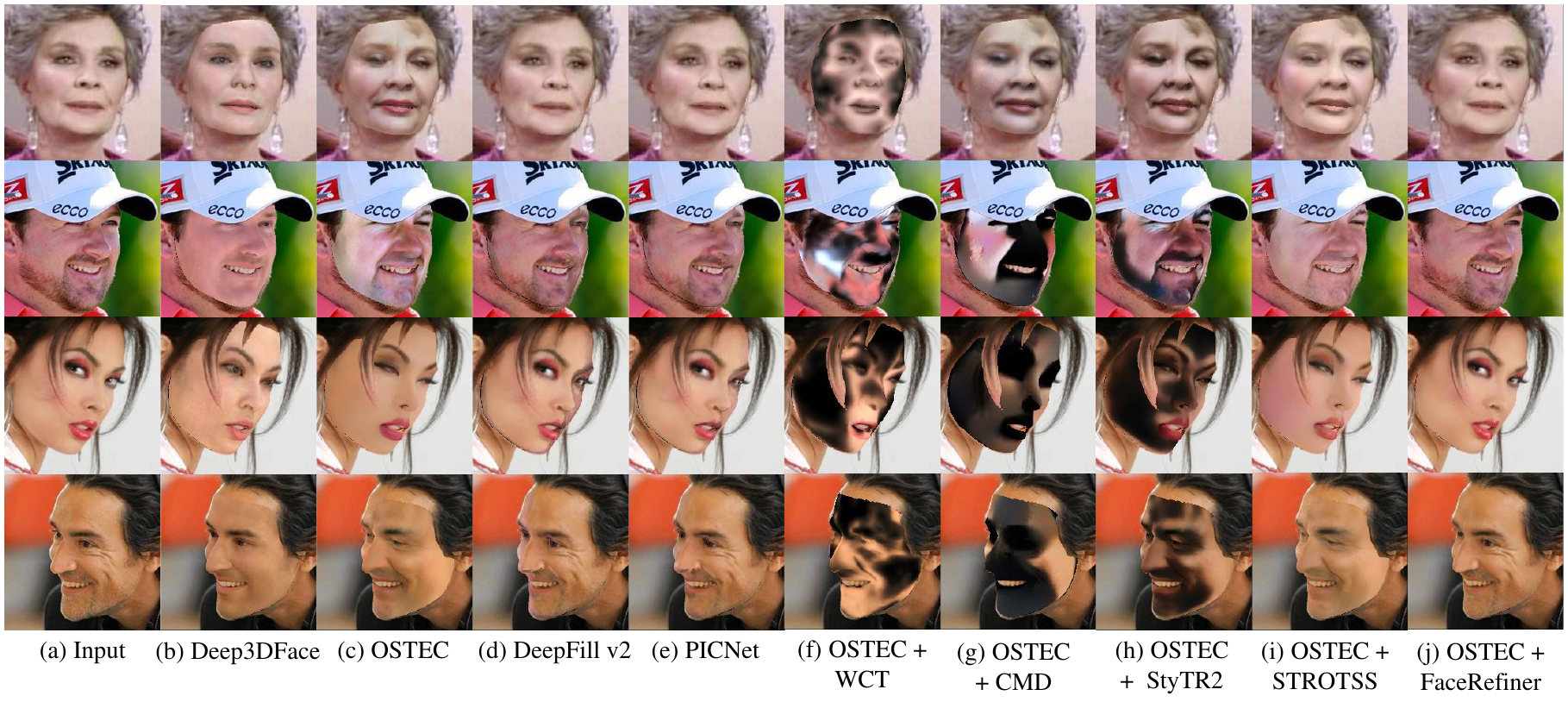} 
\caption{Qualitative comparison with the competitors on several images from CelebA. 
The generated UV textures by different methods are projected back to the input images for comparison.}
\label{fig:celebA_qualitative}
\end{figure*}

Fig. \ref{fig:coarse_content} shows the qualitative comparison between Deep3DFace and Deep3DFace + FaceRefiner on images from FFHQ \cite{styleganv1}. The facial textures generated by Deep3DFace have a uniform style, and thus can not recover the realistic face features in the regions of eyes (1st row), mouth (2nd row) and lips (3rd row). In the 2nd and 4th row, the face color is also inconsistent with inputs. The results produced by Deep3DFace come from the BFM \cite{paysan20093d} texture parametric space, which can not cover the range of in-the-wild facial textures. Fig. \ref{fig:FFHQ_qualitative} shows the qualitative comparison between OSTEC and OSTEC + FaceRefiner on the images from FFHQ. The results produced by OSTEC lose some middle and low level information such as the face spots (2nd, 4th row), face structures (1st, 3rd row) and face color (5th row). Overall, our refined results on both Deep3DFace and OSTEC contain more rich texture details and meaningful face structures. Fig. \ref{fig:celebA_qualitative} shows the qualitative evaluation results on images from CelebA. When projected back into the input images, our results can better preserve face features and face identity, compared with the competitors.

\begin{figure}
\centering
\includegraphics[width=0.46\textwidth]{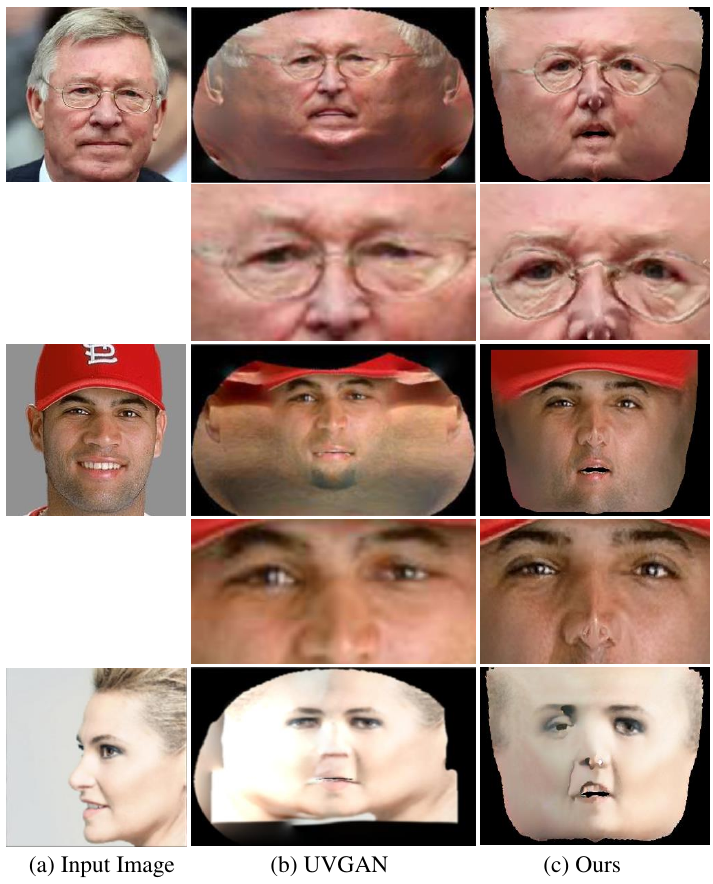} 
\caption{Comparison with UVGAN \cite{deng2018uv}. 
The images in 2nd, 4th rows are the zoom-in of images in 1st, 3rd rows.}
\label{fig:compare_uvgan}
\end{figure}

\begin{figure}
\centering
\includegraphics[width=0.46\textwidth]{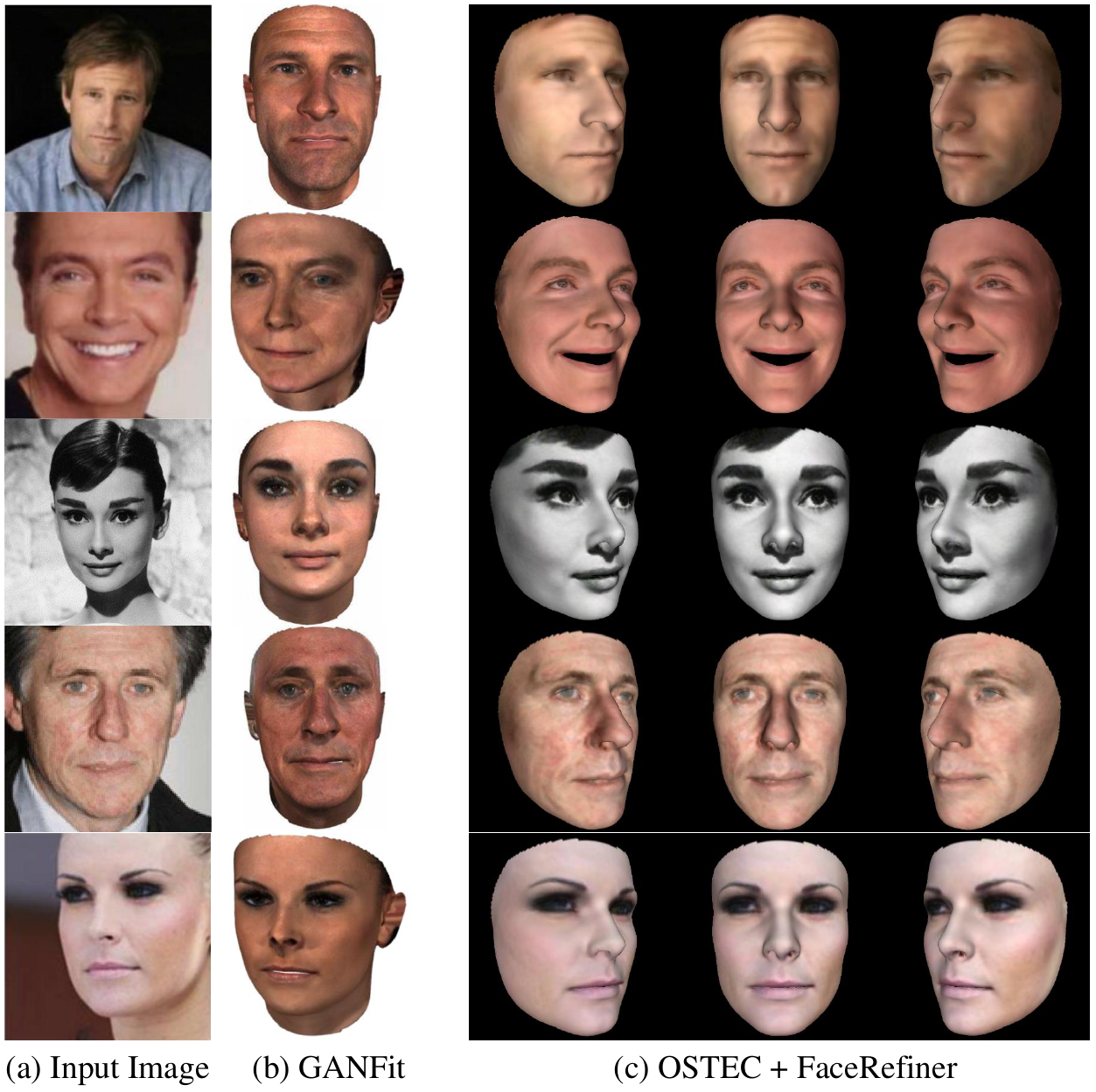} 
\caption{Comparison with GANFit \cite{gecer2019ganfit}.}
\label{fig:compare_gaifit}
\end{figure}

\begin{figure}
\centering
\includegraphics[width=0.46\textwidth]{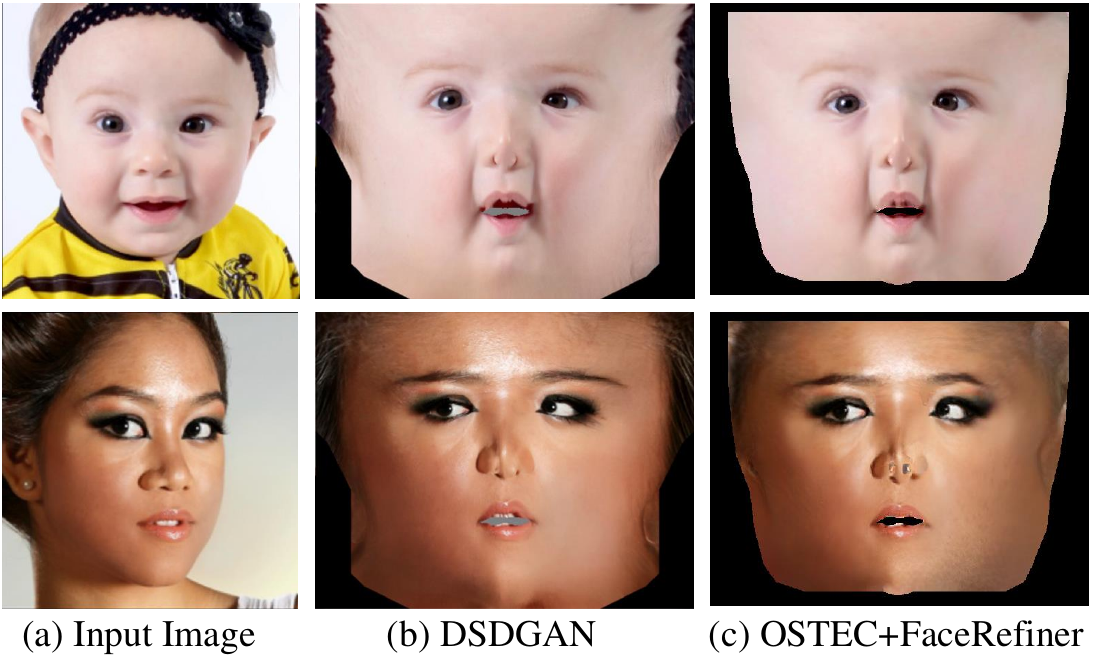} 
\caption{Comparison with DSDGAN \cite{DSDGAN2021}.}
\label{fig:compare_dsdgan}
\end{figure}

As the code, the pre-trained models and the testing dataset of UVGAN, DSDGAN and GANFit are not available online, we take the visual results from their original paper. The comparison results are shown in Fig. \ref{fig:compare_uvgan}, \ref{fig:compare_gaifit}, \ref{fig:compare_dsdgan}. The input images fed into our method in all comparisons are obtained by searching for the most similar images on the Internet and the public datasets.

\textbf{Comparison with UVGAN.} UVGAN can only generate $256\times256$ textures, while our method can infer $512\times512$ textures. Hence, as shown in 
Fig. \ref{fig:compare_uvgan}, the textures produced by UVGAN are usually more blurred than ours. Besides, UVGAN sometimes can not produce satisfactory textures for large-pose faces, e.g. face highlight in the 5th row of Fig. \ref{fig:compare_uvgan}.

\textbf{Comparison with GANFit.} As shown in Fig. \ref{fig:compare_gaifit}, although the facial textures generated by GANFit exhibit fine face details, they are drifted from the input images in face color, face identity and face expression.

\textbf{Comparison with DSDGAN.} As shown in Fig. \ref{fig:compare_dsdgan}, 
our OSTEC + FaceRefiner shows very competitive performance compared with DSDGAN. We believe that, if the code of DSDGAN is available, our DSDGAN + FaceRefiner is also effective and outperforms all other methods.

\begin{figure}
\centering
\includegraphics[width=0.48\textwidth]{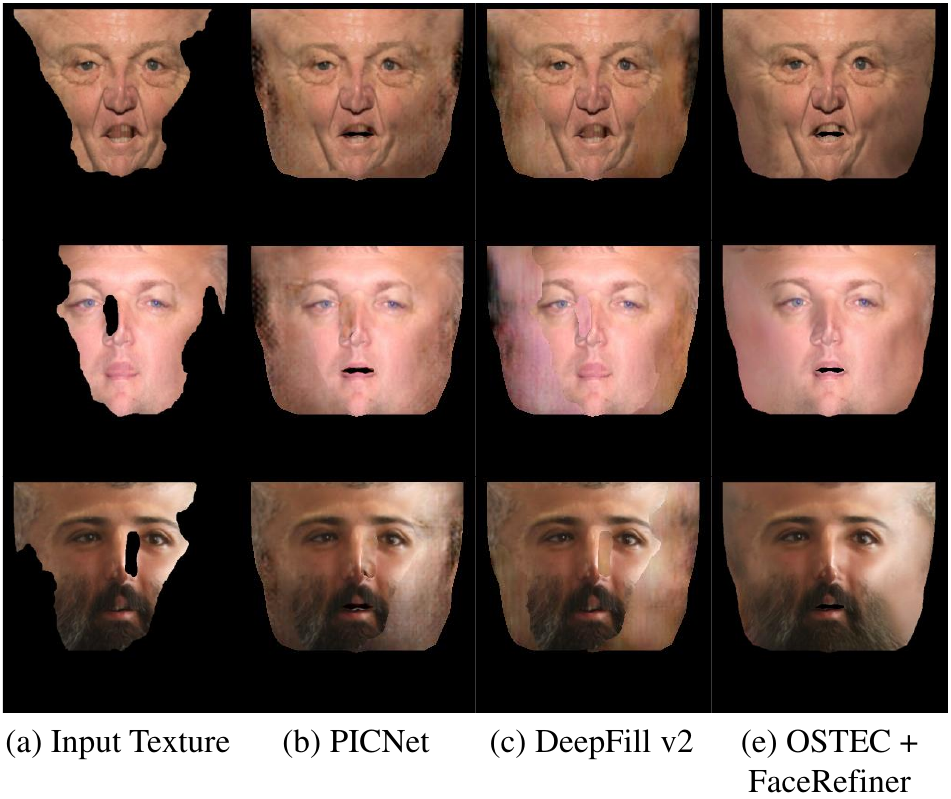} 
\caption{Qualitative comparison with image inpainting methods.}
\label{fig:image_inpainting}
\end{figure}

\subsection{Comparison with image inpainting methods}
We select two image inpainting methods as competitors: DeepFill v2 \cite{yu2019free} and PICNet \cite{zheng2019pluralistic}. We do not directly use the pre-trained models released by their authors, because they are trained on the image space (e.g. CelebA-HQ \cite{karras2017progressive}) and can not process UV textures. 
To make a fair comparison, we randomly sampled 5,000 complete textures from the texture model of BFM \cite{paysan20093d}, and then produce a set of ``masked and complete" pairs to construct the training data for the two image inpainting methods. The quantitative results in Table \ref{tab:MultiPIE} and \ref{tab:CelebA} show that our FaceRefiner + Deep3DFace or OSTEC outperforms the image inpainting methods in all metrics. 

\begin{figure}
\centering
\includegraphics[width=0.48\textwidth]{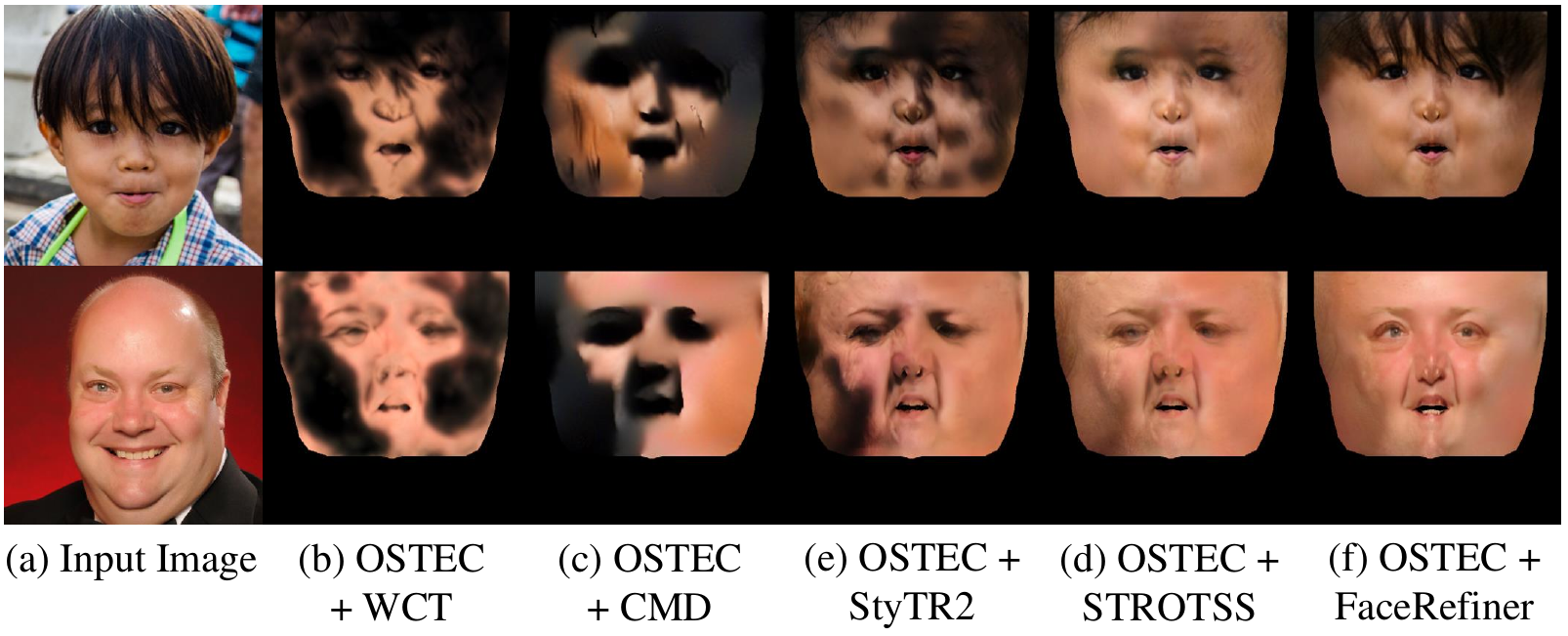} 
\caption{Qualitative comparison with other style transfer methods.}
\label{style_transfer}
\end{figure}

From Fig. \ref{fig:image_inpainting}, we can observe that although the specially trained image inpainting models are able to recover some structures and details, they can not ensure global smoothness in the UV space. There are obvious boundaries between the visible and invisible regions, since they can not well handle such large holes. The refined textures by our method are silkier than the two competitors. The same phenomenon appears in Fig. \ref{fig:celebA_qualitative}, where the reconstructed facial images by the two competitors have high quality in the visible regions, but contain artifacts (black line) around the face boundaries.

\subsection{Comparison with style transfer methods}
\label{sect:compareStyleTransfer}
As there exists many other style transfer methods, we select four style transfer methods to make comparisons, including WCT \cite{li2017universal}, CMD \cite{kalischek2021light}, StyTR2 \cite{deng2022stytr2} and STROTSS \cite{kolkin2019style}. The statistics in Table \ref{tab:MultiPIE} and \ref{tab:CelebA} clearly show that it is not effective for other style transfer methods to act as the facial texture refiner, even if the original STROTSS. Moreover, Fig. \ref{style_transfer} shows more clearly that the common style transfer methods without special designs do not yield reasonable results. The reason of the poor results produced by WCT, CMD and StyTR2 is that, there are many invalid regions (denoted in black color) in the style image, and CMD, WCT and StyTR2 migrate the style features in these regions to the result. The hypercolumn matching used in STROTSS and our FaceRefinner automatically constructs the correspondence of samples between style image and result,  thus significantly reducing the migration of black pixels to the facial regions. Although using STROTSS alone can get some reasonable UV textures, it can't migrate the identity information to the results.

%The hypercolumn-based loss function used in STROTSS and our FaceRefinner automatically matches the closest two hypercolumn, thus significantly reducing the migration of black pixels to the facial regions.

\begin{table*}
\centering
\begin{tabular}{c|cccc|cccc}
\hline
\multicolumn{1}{l|}{} & \multicolumn{4}{c|}
{PSNR $\uparrow$}                                                                                                            & \multicolumn{4}{c}{SSIM $\uparrow$}                                                                                                         \\ \hline
                      & 0°                              & ±30°                            & ±60°                            & mean                           & 0°                             & ±30°                           & ±60°                           & mean                          \\ \hline
w/o style\_loss       & 23.3600                         & 22.1647                         & 20.8966                         & 21.8965                        & 0.7966                         & 0.7857                         & 0.7680                         & 0.7808                        \\
w/o content\_loss     & 25.8751                         & 23.9999                         & 22.1480                         & 23.6342                        & 0.8863                         & 0.8694                         & 0.8557                         & 0.8673                        \\
w/o recon loss        & 23.7818                         & 22.0460                         & 20.4635                         & 21.7602                        & 0.8724                         & 0.8576                         & 0.8317                         & 0.8502                        \\
recon loss (UV space) & 26.2835                         & 24.8182                         & 22.4515                         & 24.1646                        & 0.8822                         & 0.8726                         & 0.8589                         & 0.8691                        \\ \hline
stage 1               & \cellcolor[HTML]{FFFFFF}21.8056 & \cellcolor[HTML]{FFFFFF}21.6456 & \cellcolor[HTML]{FFFFFF}20.6656 & 21.2856                        & \cellcolor[HTML]{FFFFFF}0.8697 & \cellcolor[HTML]{FFFFFF}0.8581 & \cellcolor[HTML]{FFFFFF}0.8347 & 0.8511                        \\
stage 2               & 24.3188                         & 24.1192                         & 22.4582                         & 23.4947                        & 0.8780                         & 0.8676                         & 0.8489                         & 0.8622                        \\
stage 3               & \cellcolor[HTML]{FFFFFF}26.3611 & \cellcolor[HTML]{FFFFFF}25.1945 & \cellcolor[HTML]{FFFFFF}22.9393 & 24.5257                        & \cellcolor[HTML]{FFFFFF}0.8849 & \cellcolor[HTML]{FFFFFF}0.8726 & \cellcolor[HTML]{FFFFFF}0.8553 & 0.8681                        \\
stage 4               & {\color[HTML]{3531FF} 26.8882}  & {\color[HTML]{3531FF} 25.4853}  & {\color[HTML]{FF0000} 23.0037}  & {\color[HTML]{3531FF} 24.7732} & {\color[HTML]{3531FF} 0.8873}  & {\color[HTML]{3531FF} 0.8746}  & {\color[HTML]{3531FF} 0.8585}  & {\color[HTML]{3531FF} 0.8707} \\
stage 5               & {\color[HTML]{FF0000} 27.0096}  & {\color[HTML]{FF0000} 25.5164}  & {\color[HTML]{3531FF} 22.9465}  & {\color[HTML]{FF0000} 24.7871} & {\color[HTML]{FF0000} 0.8879}  & {\color[HTML]{FF0000} 0.8753}  & {\color[HTML]{FF0000} 0.8600}  & {\color[HTML]{FF0000} 0.8717} \\ \hline

\end{tabular}
\vspace{2mm}
\caption{The quantitative results of ablation study over the Multi-PIE dataset.}
\label{Table 1}
\end{table*}

\subsection{Ablation Study}
To demonstrate the effectiveness of the novel designs in the style transfer, we conduct multiple sets of experiments in different configurations: 
(1) removing $L_{style}$, 
(2) removing $L_{content}$, 
(3) removing $L_{render}$, 
(4) replacing $L_{render}$ with the UV reconstruction loss, 
(5) increasing the number of stages sequentially from one to five. 
The first four configurations use five-stages style transfer.  
All the above experiments are conducted to refine the results produced by OSTEC.

\begin{figure}
\centering
\includegraphics[width=0.47\textwidth]{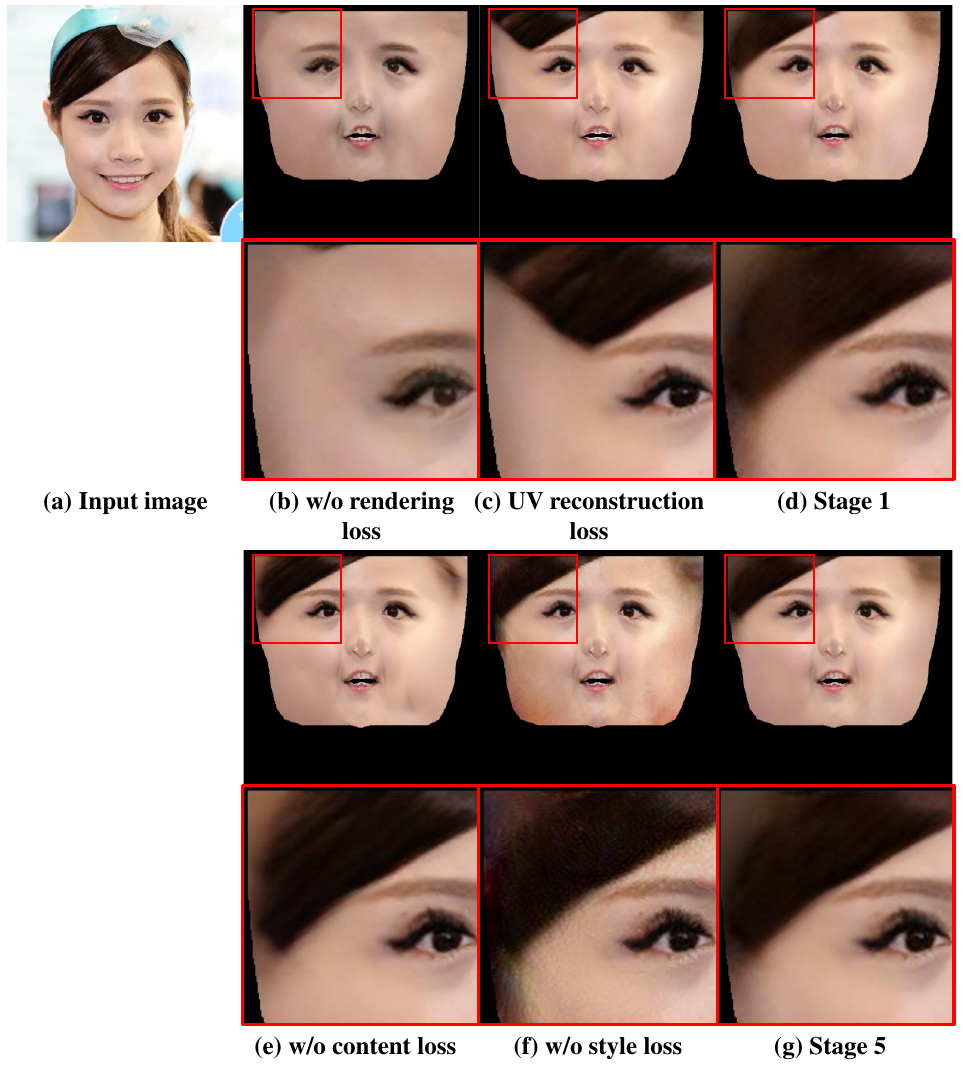} 
\caption{Qualitative comparison of FaceRefiner with different configurations in the ablation study.}
\label{fig:ablation}
\end{figure}

The testing dataset in Multi-PIE is used for evaluation, and the quantitative evaluation results are presented in Table \ref{Table 1}.
The statistics show that removing $L_{style}$, removing $L_{content}$, removing $L_{render}$ or replacing $L_{render}$ with the UV reconstruction loss decrease both PSNR and SSIM significantly. Another observation is that PSNR and SSIM increase with the number of migrations. Five-stage style transfer leads to the best results, but its metrics are not much higher than the four-stage. To balance efficiency and quality, we set the number of stages as five in all experiments.

The qualitative evaluation results are shown in Fig. \ref{fig:ablation}.
The texture in the visible parts cannot be well recovered without using $L_{render}$, as there is a big visual discrepancy between the inferred texture
(b) and the input image (a). As shown in (c), the reconstruction loss in the UV space can not work on some pixels in the hair regions. The main purpose of using multi-stage style transfer is to eliminate traces of texture migration and make the texture closer to the input image. 
As shown in (d) and (g), with the number of stages increasing, the hair color and eyes become more consistent with the input image. Removing $L_{content}$ reduces the reconstruction quality of hairs, such as the upper left and upper right corners in (e).
After removing $L_{style}$, the overall color of the result is unreasonable, and the abnormal roughness appears in the texture, as shown in (f).

\subsection{Extended Experimental Results}
\begin{figure*}
\centering
\includegraphics[width=0.96\textwidth]{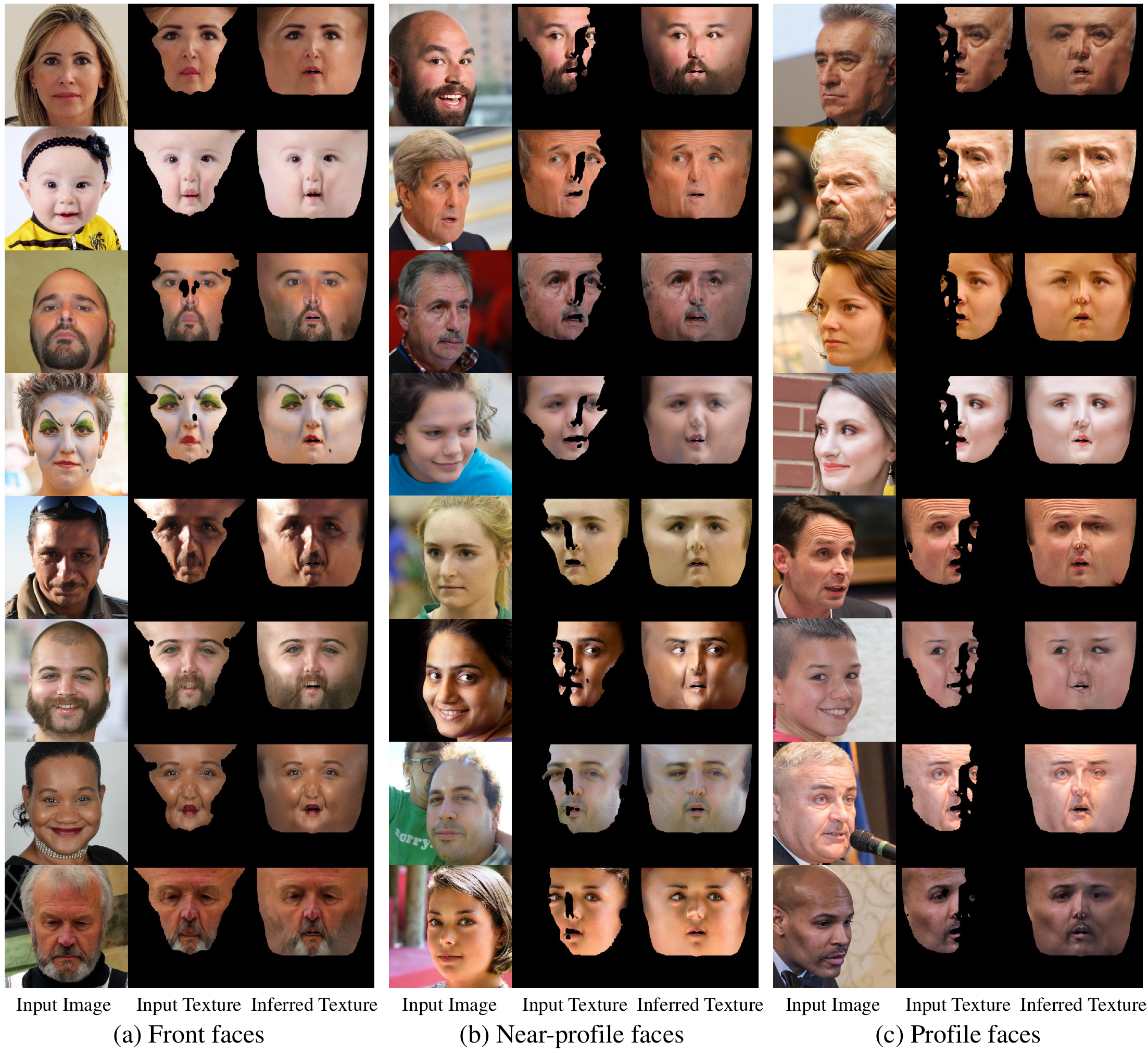} 
\caption{Extended inferred results of our method in front, near-profile and profile face images. }
\label{fig:supp_pose}
\end{figure*}

In Fig. \ref{fig:supp_pose}, we show more generated facial textures by OSTEC + FaceRefiner. The input images include front faces, near-profile faces and profile faces.

\section{conclusion and Limitations}
Current facial texture generation methods usually rely on the training on the specific dataset or the pre-trained StyleGAN. When generalizing to in-the-wild images, they entail the information loss in many aspects, such as facial details, structures and identity. Motivated by this, in this paper, we propose a novel facial texture refinement method named FaceRefiner. To realize the goal of transferring low, middle and high level information from inputs to result, we propose the differentiable rendering-based style transfer. The extensive experimental results clearly show our advantages in texture quality improving and face identity preserving. 

One limitation of FaceRefiner is that it will introduce artifacts around nose, when the input face has large pose. The reason is that it is hard to get an accurate 3D face model under large poses using existing 3D face reconstruction methods. In future, we plan to 1) reduce the time of optimization in style transfer, 2) design the multiple view style transfer method.


\ifCLASSOPTIONcaptionsoff
  \newpage
\fi

\bibliographystyle{IEEEtran}
\bibliography{IEEEexample}

@article{selim2016painting,
  title={Painting style transfer for head portraits using convolutional neural networks},
  author={Selim, Ahmed and Elgharib, Mohamed and Doyle, Linda},
  journal={ACM Transactions on Graphics},
  volume={35},
  number={4},
  pages={1--18},
  year={2016},
}

@article{champandard2016semantic,
  title={Semantic style transfer and turning two-bit doodles into fine artworks},
  author={Champandard, Alex J},
  journal={arXiv preprint arXiv:1603.01768},
  year={2016}
}

@inproceedings{li2016combining,
  title={Combining markov random fields and convolutional neural networks for image synthesis},
  author={Li, Chuan and Wand, Michael},
  booktitle={Proc. of CVPR},
  pages={2479--2486},
  year={2016}
}

@inproceedings{sanakoyeu2018style,
  title={A style-aware content loss for real-time hd style transfer},
  author={Sanakoyeu, Artsiom and Kotovenko, Dmytro and Lang, Sabine and Ommer, Bjorn},
  booktitle={Proc. of ECCV},
  pages={698--714},
  year={2018}
}

@inproceedings{johnson2016perceptual,
  title={Perceptual losses for real-time style transfer and super-resolution},
  author={Johnson, Justin and Alahi, Alexandre and Fei-Fei, Li},
  booktitle={Proc. of ECCV},
  pages={694--711},
  year={2016},
}

@inproceedings{gatys2016image,
  title={Image style transfer using convolutional neural networks},
  author={Gatys, Leon A and Ecker, Alexander S and Bethge, Matthias},
  booktitle={Proc. of CVPR},
  pages={2414--2423},
  year={2016}
}

@inproceedings{kolkin2019style,
  title={Style transfer by relaxed optimal transport and self-similarity},
  author={Kolkin, Nicholas and Salavon, Jason and Shakhnarovich, Gregory},
  booktitle={Proc. of CVPR},
  pages={10051--10060},
  year={2019}
}

@inproceedings{gecer2021ostec,
  title={Ostec: One-shot texture completion},
  author={Gecer, Baris and Deng, Jiankang and Zafeiriou, Stefanos},
  booktitle={Proc. of CVPR},
  pages={7628--7638},
  year={2021}
}

@inproceedings{yu2018generative,
  title={Generative image inpainting with contextual attention},
  author={Yu, Jiahui and Lin, Zhe and Yang, Jimei and Shen, Xiaohui and Lu, Xin and Huang, Thomas S},
  booktitle={Proc. of CVPR},
  pages={5505--5514},
  year={2018}
}

@inproceedings{liu2018image,
  title={Image inpainting for irregular holes using partial convolutions},
  author={Liu, Guilin and Reda, Fitsum A and Shih, Kevin J and Wang, Ting-Chun and Tao, Andrew and Catanzaro, Bryan},
  booktitle={Proc. of ECCV},
  pages={85--100},
  year={2018}
}

@inproceedings{yu2019free,
  title={Free-form image inpainting with gated convolution},
  author={Yu, Jiahui and Lin, Zhe and Yang, Jimei and Shen, Xiaohui and Lu, Xin and Huang, Thomas S},
  booktitle={Proc. of CVPR},
  pages={4471--4480},
  year={2019}
}

@inproceedings{deng2019accurate,
  title={Accurate 3d face reconstruction with weakly-supervised learning: From single image to image set},
  author={Deng, Yu and Yang, Jiaolong and Xu, Sicheng and Chen, Dong and Jia, Yunde and Tong, Xin},
  booktitle={Proc. of CVPRW},
  pages={285--295},
  year={2019}
}

@inproceedings{gecer2019ganfit,
  title={Ganfit: Generative adversarial network fitting for high fidelity 3d face reconstruction},
  author={Gecer, Baris and Ploumpis, Stylianos and Kotsia, Irene and Zafeiriou, Stefanos},
  booktitle={Proc. of CVPR},
  pages={1155--1164},
  year={2019}
}

@inproceedings{paysan20093d,
  title={A 3D face model for pose and illumination invariant face recognition},
  author={Paysan, Pascal and Knothe, Reinhard and Amberg, Brian and Romdhani, Sami and Vetter, Thomas},
  booktitle={International Conference on Advanced Video and Signal Based Surveillance},
  pages={296--301},
  year={2009}
}

@inproceedings{lin2020towards,
  title={Towards high-fidelity 3D face reconstruction from in-the-wild images using graph convolutional networks},
  author={Lin, Jiangke and Yuan, Yi and Shao, Tianjia and Zhou, Kun},
  booktitle={Proc. of CVPR},
  pages={5891--5900},
  year={2020}
}

@inproceedings{karras2020analyzing,
  title={Analyzing and improving the image quality of stylegan},
  author={Karras, Tero and Laine, Samuli and Aittala, Miika and Hellsten, Janne and Lehtinen, Jaakko and Aila, Timo},
  booktitle={Proc. of CVPR},
  pages={8110--8119},
  year={2020}
}

@inproceedings{hariharan2015hypercolumns,
  title={Hypercolumns for object segmentation and fine-grained localization},
  author={Hariharan, Bharath and Arbel{\'a}ez, Pablo and Girshick, Ross and Malik, Jitendra},
  booktitle={Proc. of CVPR},
  pages={447--456},
  year={2015}
}

@inproceedings{mostajabi2015feedforward,
  title={Feedforward semantic segmentation with zoom-out features},
  author={Mostajabi, Mohammadreza and Yadollahpour, Payman and Shakhnarovich, Gregory},
  booktitle={Proc. of CVPR},
  pages={3376--3385},
  year={2015}
}

@article{simonyan2014very,
  title={Very deep convolutional networks for large-scale image recognition},
  author={Simonyan, Karen and Zisserman, Andrew},
  journal={arXiv preprint arXiv:1409.1556},
  year={2014}
}

@inproceedings{kim2020deformable,
  title={Deformable style transfer},
  author={Kim, Sunnie SY and Kolkin, Nicholas and Salavon, Jason and Shakhnarovich, Gregory},
  booktitle={Proc. of ECCV},
  pages={246--261},
  year={2020}
}

@article{gross2010multi,
  title={Multi-PIE},
  author={Gross, Ralph and Matthews, Iain and Cohn, Jeffrey and Kanade, Takeo and Baker, Simon},
  journal={Image and Vision Computing},
  volume={28},
  number={5},
  pages={807--813},
  year={2010}
}

@inproceedings{zheng2019pluralistic,
  title={Pluralistic image completion},
  author={Zheng, Chuanxia and Cham, Tat-Jen and Cai, Jianfei},
  booktitle={Proc. of CVPR},
  pages={1438--1447},
  year={2019}
}

@inproceedings{karras2017progressive,
  author       = {Tero Karras and
                  Timo Aila and
                  Samuli Laine and
                  Jaakko Lehtinen},
  title        = {Progressive Growing of GANs for Improved Quality, Stability, and Variation},
  booktitle    = {Proc. of ICLR},
  year         = {2018}
}

@inproceedings{chen2022towards,
  title={Towards High-Fidelity Face Self-occlusion Recovery via Multi-view Residual-based GAN Inversion},
  author={Chen, Jinsong and Han, Hu and Shan, Shiguang},
  booktitle={Proc. of AAAI},
  pages = {294--302},
  year={2022}
}

@inproceedings{alaluf2021restyle,
  title={Restyle: A residual-based stylegan encoder via iterative refinement},
  author={Alaluf, Yuval and Patashnik, Or and Cohen-Or, Daniel},
  booktitle={Proc. of CVPR},
  pages={6711--6720},
  year={2021}
}

@inproceedings{liu2021deep,
  title={Deep Style Transfer for Line Drawings},
  author={Liu, Xueting and Wu, Wenliang and Wu, Huisi and Wen, Zhenkun},
  booktitle={Proc. of AAAI},
  volume={35},
  number={1},
  pages={353--361},
  year={2021}
}

@inproceedings{yang2020learning,
  title={Learning to incorporate structure knowledge for image inpainting},
  author={Yang, Jie and Qi, Zhiquan and Shi, Yong},
  booktitle={Proc. of AAAI},
  pages={12605--12612},
  year={2020}
}

@inproceedings{hertzmann2001image,
  title={Image analogies},
  author={Hertzmann, Aaron and Jacobs, Charles E and Oliver, Nuria and Curless, Brian and Salesin, David H},
  booktitle={Proc. of SIGGRAPH},
  pages={327--340},
  year={2001}
}

@article{shih2013data,
  title={Data-driven hallucination of different times of day from a single outdoor photo},
  author={Shih, Yichang and Paris, Sylvain and Durand, Fr{\'e}do and Freeman, William T},
  journal={ACM Transactions on Graphics},
  volume={32},
  number={6},
  pages={1--11},
  year={2013}
}

@inproceedings{li2020fet,
  title={FET-GAN: Font and effect transfer via k-shot adaptive instance normalization},
  author={Li, Wei and He, Yongxing and Qi, Yanwei and Li, Zejian and Tang, Yongchuan},
  booktitle={Proc. of AAAI},
  pages={1717--1724},
  year={2020}
}

@inproceedings{deng2021arbitrary,
  title={Arbitrary video style transfer via multi-channel correlation},
  author={Deng, Yingying and Tang, Fan and Dong, Weiming and Huang, Haibin and Ma, Chongyang and Xu, Changsheng},
  booktitle={Proc. of AAAI},
  pages={1210--1217},
  year={2021}
}

@inproceedings{deng2018uv,
  title={Uv-gan: Adversarial facial uv map completion for pose-invariant face recognition},
  author={Deng, Jiankang and Cheng, Shiyang and Xue, Niannan and Zhou, Yuxiang and Zafeiriou, Stefanos},
  booktitle={Proc. of CVPR},
  pages={7093--7102},
  year={2018}
}

@inproceedings{lin2021meingame,
  title={Meingame: Create a game character face from a single portrait},
  author={Lin, Jiangke and Yuan, Yi and Zou, Zhengxia},
  booktitle={Proc. of AAAI},
  pages={311--319},
  year={2021}
}

@inproceedings{booth20173d,
  title={3d face morphable models in-the-wild},
  author={Booth, James and Antonakos, Epameinondas and Ploumpis, Stylianos and Trigeorgis, George and Panagakis, Yannis and Zafeiriou, Stefanos},
  booktitle={Proc. of CVPR},
  pages={48--57},
  year={2017}
}

@inproceedings{DSDGAN2021,
  author    = {Jongyoo Kim and
               Jiaolong Yang and
               Xin Tong},
  title     = {Learning High-Fidelity Face Texture Completion without Complete Face               Texture},
  booktitle = {Proc. of ICCV},
  pages     = {13970--13979},
  year      = {2021},
}

@inproceedings{Hassner_2015_CVPR,
  author = {Hassner, Tal and Harel, Shai and Paz, Eran and Enbar, Roee},
  title = {Effective Face Frontalization in Unconstrained Images},
  booktitle = {Proc. of CVPR},
  pages = {4295--4304},
  year = {2015}
}

@article{ichim2015dynamic,
  title={Dynamic 3D avatar creation from hand-held video input},
  author={Ichim, Alexandru Eugen and Bouaziz, Sofien and Pauly, Mark},
  journal={ACM Transactions on Graphics},
  volume={34},
  number={4},
  pages={1--14},
  year={2015},
}

@inproceedings{lattas2020avatarme,
  title={AvatarMe: Realistically Renderable 3D Facial Reconstruction" in-the-wild"},
  author={Lattas, Alexandros and Moschoglou, Stylianos and Gecer, Baris and Ploumpis, Stylianos and Triantafyllou, Vasileios and Ghosh, Abhijeet and Zafeiriou, Stefanos},
  booktitle={Proc. of CVPR},
  pages={760--769},
  year={2020}
}

@inproceedings{ploumpis2019combining,
  title={Combining 3d morphable models: A large scale face-and-head model},
  author={Ploumpis, Stylianos and Wang, Haoyang and Pears, Nick and Smith, William AP and Zafeiriou, Stefanos},
  booktitle={Proc. of CVPR},
  pages={10934--10943},
  year={2019}
}

@inproceedings{blanz1999morphable,
  title={A morphable model for the synthesis of 3D faces},
  author={Blanz, Volker and Vetter, Thomas},
  booktitle={Proc. of SIGGRAPH},
  pages={187--194},
  year={1999}
}

@article{wang2004image,
  title={Image quality assessment: from error visibility to structural similarity},
  author={Wang, Zhou and Bovik, Alan C and Sheikh, Hamid R and Simoncelli, Eero P},
  journal={IEEE Transactions on Image Processing},
  volume={13},
  number={4},
  pages={600--612},
  year={2004},
  publisher={IEEE}
}

@article{wu2018light,
  title={A light CNN for deep face representation with noisy labels},
  author={Wu, Xiang and He, Ran and Sun, Zhenan and Tan, Tieniu},
  journal={IEEE Transactions on Information Forensics and Security},
  volume={13},
  number={11},
  pages={2884--2896},
  year={2018}
}

@article{zhao2019multi,
  title={Multi-prototype networks for unconstrained set-based face recognition},
  author={Zhao, Jian and Li, Jianshu and Tu, Xiaoguang and Zhao, Fang and Xin, Yuan and Xing, Junliang and Liu, Hengzhu and Yan, Shuicheng and Feng, Jiashi},
  journal={arXiv preprint arXiv:1902.04755},
  year={2019}
}

@inproceedings{liuICCV2015,
  author       = {Ziwei Liu and
                  Ping Luo and
                  Xiaogang Wang and
                  Xiaoou Tang},
  title        = {Deep Learning Face Attributes in the Wild},
  booktitle    = {Proc. of ICCV},
  pages        = {3730--3738},
  year         = {2015},
}

@inproceedings{li2017universal,
  title        = {Universal Style Transfer via Feature Transforms},
  author       = {Yijun Li and
                  Chen Fang and
                  Jimei Yang and
                  Zhaowen Wang and
                  Xin Lu and
                  Ming{-}Hsuan Yang},
  booktitle    = {Proc. of NIPS},
  pages        = {386--396},
  year         = {2017}
}

@inproceedings{kalischek2021light,
  title={In the light of feature distributions: moment matching for neural style transfer},
  author={Kalischek, Nikolai and Wegner, Jan D and Schindler, Konrad},
  booktitle={Proc. of CVPR},
  pages={9382--9391},
  year={2021}
}

@inproceedings{styleganv1,
  author       = {Tero Karras and
                  Samuli Laine and
                  Timo Aila},
  title        = {A Style-Based Generator Architecture for Generative Adversarial Networks},
  booktitle    = {Proc. of CVPR},
  pages        = {4401--4410},
  year         = {2019}
}

@article{StripsST,
  author       = {Yujie Huang and
                  Yuhao Liu and
                  Ming{-}e Jing and
                  Xiaoyang Zeng and
                  Yibo Fan},
  title        = {Tear the Image Into Strips for Style Transfer},
  journal      = {IEEE Transactions on Multimedia},
  volume       = {24},
  pages        = {3978--3988},
  year         = {2022}
}

@article{VideoST2021,
  author       = {Kai Xu and
                  Longyin Wen and
                  Guorong Li and
                  Honggang Qi and
                  Liefeng Bo and
                  Qingming Huang},
  title        = {Learning Self-Supervised Space-Time {CNN} for Fast Video Style Transfer},
  journal      = {IEEE Transactions on Multimedia},
  volume       = {30},
  pages        = {2501--2512},
  year         = {2021}
}

@article{DCT2023,
  author={Liu, Renshuai and Cheng, Yao and Huang, Sifei and Li, Chengyang and Cheng, Xuan},
  journal={IEEE Transactions on Multimedia}, 
  title={Transformer-based High-Fidelity Facial Displacement Completion for Detailed 3D Face Reconstruction}, 
  year={2023},
  volume={},
  number={},
  pages={1-13},
  doi={10.1109/TMM.2023.3271816}
}

@article{TexturePreservingST2022,
  author       = {Hwanbok Mun and
                  Gang{-}Joon Yoon and
                  Jinjoo Song and
                  Sang Min Yoon},
  title        = {Texture Preserving Photo Style Transfer Network},
  journal      = {IEEE Transactions on Multimedia},
  volume       = {24},
  pages        = {3823--3834},
  year         = {2022}
}

@article{laine2020modular,
  title={Modular primitives for high-performance differentiable rendering},
  author={Laine, Samuli and Hellsten, Janne and Karras, Tero and Seol, Yeongho and Lehtinen, Jaakko and Aila, Timo},
  journal={ACM Transactions on Graphics},
  volume={39},
  number={6},
  pages={1--14},
  year={2020},
  publisher={ACM New York, NY, USA}
}

@article{kong2023exploring,
  title={Exploring the temporal consistency of arbitrary style transfer: A channelwise perspective},
  author={Kong, Xiaoyu and Deng, Yingying and Tang, Fan and Dong, Weiming and Ma, Chongyang and Chen, Yongyong and He, Zhenyu and Xu, Changsheng},
  journal={IEEE Transactions on Neural Networks and Learning Systems},
  volume={},
  number={},
  pages={1-15},
  year={2023}
}

@inproceedings{deng2022stytr2,
  title={Stytr2: Image style transfer with transformers},
  author={Deng, Yingying and Tang, Fan and Dong, Weiming and Ma, Chongyang and Pan, Xingjia and Wang, Lei and Xu, Changsheng},
  booktitle={Proc. of CVPR},
  pages={11326--11336},
  year={2022}
}

@inproceedings{psnr2010,
  author       = {Alain Hor{\'{e}} and
                  Djemel Ziou},
  title        = {Image Quality Metrics: {PSNR} vs. {SSIM}},
  booktitle    = {Proc. of ICPR},
  pages        = {2366--2369},
  year         = {2010},
}

@inproceedings{Artflow2021,
  author       = {Jie An and
                  Siyu Huang and
                  Yibing Song and
                  Dejing Dou and
                  Wei Liu and
                  Jiebo Luo},
  title        = {ArtFlow: Unbiased Image Style Transfer via Reversible Neural Flows},
  booktitle    = {Proc. of CVPR},
  pages        = {862--871},
  year         = {2021}
}

@inproceedings{emef2023,
  author       = {Renshuai Liu and
                  Chengyang Li and
                  Haitao Cao and
                  Yinglin Zheng and
                  Ming Zeng and
                  Xuan Cheng},
  title        = {{EMEF:} Ensemble Multi-Exposure Image Fusion},
  booktitle    = {Proc. of AAAI},
  pages        = {1710--1718},
  year         = {2023}
}

@inproceedings{DNPM,
  author       = {Haitao Cao and
                  Baoping Cheng and
                  Qiran Pu and
                  Haocheng Zhang and
                  Bin Luo and
                  Yixiang Zhuang and
                  Juncong Lin and
                  Liyan Chen and
                  Xuan Cheng},
  title        = {{DNPM:} {A} Neural Parametric Model for the Synthesis of Facial Geometric
                  Details},
  booktitle    = {Proc. of ICME},
  pages        = {1--6},
  year         = {2024},
}

\end{document}